\documentclass[5p, review]{elsarticle}
\makeatletter 
\def\ps@pprintTitle{%
 \let\@oddhead\@empty
 \let\@evenhead\@empty
 \let\@evenfoot\@oddfoot}
\makeatother
\usepackage[utf8]{inputenc}
\usepackage[T1]{fontenc}
\usepackage[english]{babel}
\usepackage[babel=true]{csquotes}
\usepackage[fleqn]{amsmath} 
\usepackage{amsthm} 
\usepackage{booktabs}
\usepackage{multirow} 
\usepackage{amssymb} 
\usepackage{float}
\usepackage{hyperref} 
\usepackage[english]{cleveref} 
\usepackage{subcaption}
\usepackage{caption}
\usepackage{natbib}
\usepackage{soul}
\usepackage{color}
\usepackage[dvipsnames,table]{xcolor}
\usepackage{pifont}
\usepackage{graphicx}
\usepackage{svg}
\usepackage[inline]{enumitem}
\usepackage{array}
\usepackage{longtable}
\usepackage{tabularx}
\usepackage{graphicx}
\usepackage{makecell}
\usepackage{lscape}
\usepackage{threeparttable}
\usepackage{xurl}
\usepackage{wasysym}

\usepackage{algorithm}
\usepackage{algpseudocode}

\usepackage[
  separate-uncertainty = true,
  multi-part-units = repeat
]{siunitx}

\usepackage{tablefootnote}
\usepackage{footnote}
\makesavenoteenv{tabular}
\makesavenoteenv{table}
\makesavenoteenv{table*}

\setstcolor{blue}

\newtoggle{finalPaper}
\toggletrue{finalPaper} 

\iftoggle{finalPaper} {
    \newcommand{\addtxt}[1]{#1}
    
    \newcommand{\change}[2]{#2}
    \newcommand{\rmvtxt}[1]{}
} {
    \newcommand{\addtxt}[1]{\textcolor{blue}{#1}}
    
    \newcommand{\change}[2]{\st{#1}\textcolor{blue}{#2}}
    \newcommand{\rmvtxt}[1]{\st{#1}}
}

\definecolor{ao}{rgb}{0.0, 0.5, 0.0}
\definecolor{amber}{rgb}{1.0, 0.49, 0.0}
\newcommand{\yestick}{{\color{ao}\ding{51}}}
\newcommand{\notick}{{\color{red}\ding{55}}}
\newcommand{\maybetick}{{\color{amber}$\approx$}}

\definecolor{gray45}{gray}{.45}
\definecolor{gray75}{gray}{.75}
\definecolor{orange-fig}{HTML}{C55A11}

\newcommand{\figurenames}{Figures}

\journal{Expert Systems with Applications}

\biboptions{authoryear}

\begin{document}

\begin{frontmatter}

\title{Fedstellar: A Platform for Decentralized Federated Learning}

\author[1]{Enrique Tom\'as Mart\'inez Beltr\'an\corref{cor1}}\ead{enriquetomas@um.es}
\author[2]{\'Angel Luis Perales G\'omez}\ead{angelluis.perales@um.es}
\author[3]{Chao Feng}\ead{cfeng@ifi.uzh.ch}
\author[1]{Pedro Miguel S\'anchez S\'anchez}\ead{pedromiguel.sanchez@um.es}
\author[1]{Sergio L\'opez Bernal}\ead{slopez@um.es}
\author[4]{G\'er\^ome Bovet}\ead{gerome.bovet@armasuisse.ch}
\author[1]{Manuel Gil P\'erez}\ead{mgilperez@um.es}
\author[1]{Gregorio Mart\'inez P\'erez}\ead{gregorio@um.es}
\author[3]{Alberto Huertas Celdr\'an}\ead{huertas@ifi.uzh.ch}

\cortext[cor1]{Corresponding author.}

\address[1]{Department of Information and Communications Engineering, University of Murcia, 30100 Murcia, Spain}
\address[2]{Department of Computers Engineering and Technology, University of Murcia, 30100 Murcia, Spain}
\address[3]{Communication Systems Group CSG, Department of Informatics IfI, University of Zurich UZH, CH--8050 Zürich, Switzerland}
\address[4]{Cyber-Defence Campus, Armasuisse Science and Technology, 3602 Thun, Switzerland}

\begin{abstract}

In 2016, Google proposed Federated Learning (FL) as a novel paradigm to train Machine Learning (ML) models across the participants of a federation while preserving data privacy. Since its birth, Centralized FL (CFL) has been the most used approach, where a central entity aggregates participants' models to create a global one. However, CFL presents limitations such as communication bottlenecks, single point of failure, and reliance on a central server. Decentralized Federated Learning (DFL) addresses these issues by enabling decentralized model aggregation and minimizing dependency on a central entity. Despite these advances, current platforms training DFL models struggle with key issues such as managing heterogeneous federation network topologies, adapting the FL process to virtualized or physical deployments, and using a limited number of metrics to evaluate different federation scenarios for efficient implementation. To overcome these challenges, this paper presents Fedstellar, \change{a novel platform}{a platform extended from p2pfl library and} designed to train FL models in a decentralized, semi-decentralized, and centralized fashion across diverse federations of physical or virtualized devices. Fedstellar allows users to create federations by customizing parameters like the number and type of devices training FL models, the network topology connecting them, the machine and deep learning algorithms, or the datasets of each participant, among others. Additionally, it offers real-time monitoring of model and network performance. The Fedstellar implementation encompasses a web application with an interactive graphical interface, a controller for deploying federations of nodes using physical or virtual devices, and a core deployed on each device, which provides the logic needed to train, aggregate, and communicate in the network. The effectiveness of the platform has been demonstrated in two scenarios: a physical deployment involving single-board devices such as Raspberry Pis for detecting cyberattacks and a virtualized deployment comparing various FL approaches in a controlled environment using MNIST and CIFAR-10 datasets. In both scenarios, Fedstellar demonstrated consistent performance and adaptability, achieving $F_{1} \ scores$ of 91\%, 98\%, and 91.2\% using DFL for detecting cyberattacks and classifying MNIST and CIFAR-10, respectively, reducing training time by 32\% compared to centralized approaches.

\end{abstract}


\end{frontmatter}

\section{Introduction}\label{sec:introduction}

In the digital era, a remarkable proliferation of Internet of Things (IoT) devices is transforming various sectors, ranging from wearable technologies and connected vehicles to smart homes and cities. The rapid evolution of these devices has resulted in an enormous volume, variety, and speed of data generated by IoT devices, amplifying the demand for advanced computation techniques \citep{idc:iot_number:2022}. With their heavy reliance on centralized data collection and processing, traditional Machine Learning (ML) and Deep Learning (DL) techniques increasingly face privacy, scalability, computational efficiency, and latency issues \citep{Paleyes:ml-challenges:2022}.

Federated Learning (FL), an emerging and promising Artificial Intelligence (AI) paradigm, offers a potential solution to these limitations \citep{McMahan:FL_google:2016}. It facilitates the training of models across numerous devices or nodes, preserving data privacy by conducting computations locally on each device, thus avoiding the need to share raw data directly. Furthermore, FL promotes a distributed learning process across the federation, mitigating some of the primary limitations of traditional ML, such as scalability and the requirement for central data repositories. While most current tools, frameworks, and platforms for training FL models are based on Centralized Federated Learning (CFL), wherein a single participant receives models from others and performs the aggregation, this centralization introduces potential issues. These include a single point of failure and communication bottlenecks, which can negatively impact system performance and reliability \citep{Hard:google_centralized_to_dfl:2021}.

\begin{figure}[!htb]
\includegraphics[width=\columnwidth]{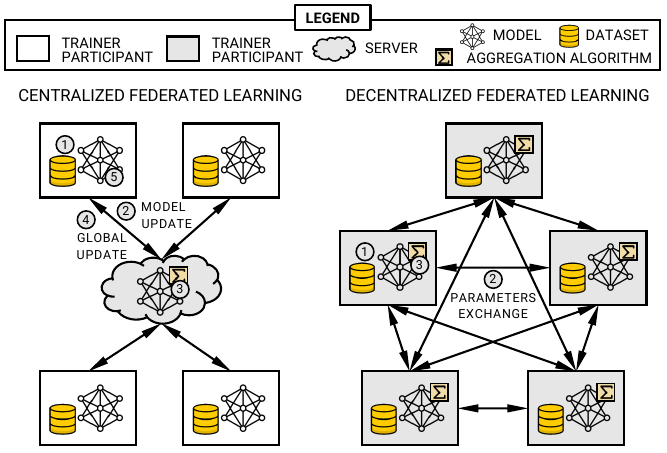}
\centering
\caption{Model training process in CFL and DFL}
\label{fig:dfl-cfl}
\end{figure}

Decentralized Federated Learning (DFL) emerged to overcome these limitations. DFL enhances decentralization by facilitating model aggregation at multiple nodes, drastically reducing dependence on a single central server \citep{MartinezBeltran:DFL_survey:2023}. This innovative structure promises to improve scalability, robustness, and efficiency, aligning seamlessly with the dynamic and distributed nature of IoT applications and enabling compatibility with emerging processors like RISC-V \citep{Wang:dfl_edge:2022, Mittone:riscv_dfl:2023}. As shown in Figure \ref{fig:dfl-cfl}, CFL and DFL vary in several steps of the model training process. Specifically, in CFL, participants train models locally (1) and forward these updates to the central server (2). The server then aggregates these updates to produce a global model (3), which is sent back to the participants (4). Finally, participants integrate this global update into their local models (5). Conversely, in DFL, participants train models locally (1), then directly exchange model parameters amongst themselves (2), and finally refine their local models by aggregating these received parameters (3), thereby highlighting the unique characteristics of their federations.

Despite the promising attributes of DFL, its adoption and practical application pose significant challenges. Over the past years, several DFL frameworks have emerged, each aiming to address the intricacies of decentralized model training. However, a closer examination reveals a multitude of recurring issues. Most current DFL platforms provide inadequate support for a variety of aggregation algorithms. Additionally, many platforms cannot handle a broad spectrum of models and datasets, often limiting researchers and practitioners to specific architectures or data formats \citep{Mittone:model_agnostic:2023}. As DFL evolves, handling various network topologies becomes paramount. Existing platforms struggle with diverse or dynamic federation structures, along with constraints in asynchronous message exchange during federation \citep{Wang:edge_communication_optimization:2021}. Furthermore, several DFL platforms are tailored for specific applications, lacking adaptability for various scenarios. Lastly, a tangible gap exists in deployment and provision. While real-world applications of DFL often involve physical devices, many platforms prioritize virtual environments, neglecting the nuances of actual device deployment.

To address these challenges and contribute to the advancement of DFL, this paper presents the following contributions:

\begin{itemize}
    \item The design and implementation of \textit{Fedstellar}, a comprehensive platform geared towards training FL models in a decentralized fashion (source code publicly available in \citep{MartinezBeltran:fedstellar_git:2023}). \addtxt{The platform employs the communication schema and models within P2P networks presented in \citep{Guijas:p2pfl_report:2022}, laying the groundwork for enhanced collaborative model training across distributed environments. At the same time, }Fedstellar supports the establishment of federations comprising diverse devices, network topologies, and algorithms. It also provides sophisticated federation management tools and performance metrics to facilitate efficient learning process monitoring. This is achieved through extensible modules that offer data storage and asynchronous capabilities alongside efficient mechanisms for model training, communication, and comprehensive analysis for federation monitoring. The platform incorporates a modular architecture comprising a user-friendly frontend for experiment setup and monitoring, a controller for effective operations orchestration, and a core component deployed in each device for model training and communication.
    
    \item The deployment of Fedstellar on two distinct federated scenarios: \textit{(i)} a physical scenario composed of five Raspberry Pi 4 and three Rock64 boards affected by cyberattacks that need to be detected, and \textit{(ii)} a virtualized scenario deployed on a server where twenty nodes used the MNIST and CIFAR-10 datasets to perform image classification tasks in a decentralized manner. In the first scenario, a fully connected topology was employed with DFL and FedAvg, using an autoencoder as the federated model. In the second scenario, different FL architectures have been benchmarked within Fedstellar, including DFL, Semi-Decentralized Federated Learning (SDFL), and CFL, under a controlled and reproducible setting. Different network topologies such as fully connected, star, and ring were employed for this comparison.
    
    \item The evaluation of the platform performance in the previous scenarios. For this purpose, various Key Performance Indicators (KPIs) were employed, including model $F_{1} \ score$, training time, communication overhead, and resource usage. In the physical scenario, Fedstellar achieved an $F_{1} \ score$ of 91\% in detecting cyberattacks using DFL with a fully connected topology. In the virtualized scenario, Fedstellar obtained an $F_{1} \ score$ of 98\% using DFL and 97.3\% using SDFL with the MNIST dataset \citep{Deng:MNIST:2012} while reducing the training time for model convergence by 32\% compared to centralized architectures. Furthermore, the effectiveness of the platform was emphasized by its adaptability to various network structures, including fully connected, star, and ring topologies.
\end{itemize}

The remainder of the paper is structured as follows. Section~\ref{sec:relatedwork} reviews the related work, discussing existing platforms and frameworks for DFL, SDFL, and CFL. After that, Section~\ref{sec:platform} describes the design and implementation of the Fedstellar platform. Section~\ref{sec:validation} validates the solution with a physical and virtualized deployment following a well-defined protocol. Subsequently, Section~\ref{sec:results} details the experiments and performance of the Fedstellar with different settings. Finally, Section \ref{sec:conclusion} presents some conclusions and future work.
\section{Related Work}\label{sec:relatedwork}

As the landscape of FL continuously evolves, numerous platforms and frameworks have emerged to adapt to diverse scenarios, addressing intricate challenges like decentralization, network topologies, scalability, and extensibility. \tablename~\ref{tab:compare-funcionality} summarizes some notable works, marking significant advances in the field.

\begin{table*}[ht!]
\caption{Comparison of functionality provided by literature solutions} \label{tab:compare-funcionality}
\resizebox{\textwidth}{!}{
\centering
\begin{threeparttable}
\begin{tabular}{c c c c c c c c c c c c}
\hline
\textbf{Solution} & \textbf{FL Architecture} & \textbf{Topology} & \textbf{Scalability} & \textbf{Extensibility} & \textbf{Failure Resilience} & \textbf{Security} & \textbf{Libraries} & \textbf{Deployment} & \textbf{Application} & \textbf{Metrics} & \textbf{Open Source} \\
\hline
\makecell{TFF\\\citep{google:tff:2019}} & CFL & Star & \maybetick & \maybetick & \notick & \makecell{Multi-party computation\\HE, DP} & TensorFlow & Simulated & Smart grid systems & \makecell{Federated models\\Computation time} & \yestick \\
\hline
\makecell{FedML\\\citep{He:fedml:2020}} & CFL & Star & \yestick & \maybetick & \yestick & \makecell{Secure aggregation} & TensorFlow, PyTorch & Hybrid & Mobile services & \makecell{Federated models\\Convergence rate} & \yestick \\
\hline
\makecell{FATE\\\citep{fedai:fate:2021}} & CFL & Star & \maybetick & \notick & \maybetick & \makecell{Multi-party computation\\HE} & \makecell{TensorFlow, PyTorch,\\C++, Java} & Simulated & \makecell{Finance\\ Credit Scoring} & Federated models & \yestick \\
\hline
\makecell{BrainTorrent\\\citep{Roy:braintorrent:2019}} & DFL & Fully connected & \textit{N/S} & \textit{N/S} & \notick & \notick & \textit{N/S} & Simulated & Healthcare & \makecell{Federated models\\CPU usage} & \notick \\
\hline
\makecell{Scatterbrained\\\citep{Wilt:scatterbrained:2021}} & DFL & Fully connected & \notick & \yestick & \notick & \notick & \makecell{PyTorch,\\Synchronous sockets} & Simulated & Academic research & Federated models & \yestick \\
\hline
\makecell{\addtxt{p2pfl}\\\citep{Guijas:p2pfl_report:2022}} & \addtxt{DFL} & \addtxt{Custom} & \maybetick & \yestick & \maybetick & \makecell{\addtxt{Communication encryption}} & \makecell{\addtxt{PyTorch, Sockets}} & \addtxt{Hybrid} & \addtxt{N/S} & \addtxt{Federated models} & \yestick \\
\hline
\makecell{FL-SEC\\\citep{Qu:fl-sec:2022}} & DFL & Fully connected & \notick & \maybetick & \notick & \makecell{Blockchain} & PyTorch & Simulated & AIoT & \makecell{Federated models} & \notick \\
\hline
\makecell{P4L\\\citep{Arapakis:P4L:2023}} & DFL & Fully connected & \yestick & \maybetick & \textit{N/S} & \makecell{Partial HE} & PyTorch & Simulated & \textit{N/S} & \makecell{Federated models\\Privacy leakage level} & \notick \\
\hline
\makecell{2DF-IDS\\\citep{Friha:2df-ids:2023}} & DFL & \makecell{Random, Ring} & \maybetick & \maybetick & \maybetick & \makecell{Communication encryption} & \makecell{Scikit-learn, Opacus,\\PyTorch} & Simulated & IIoT & \makecell{Federated models} & \notick \\
\hline
\makecell{DEFEAT\\\citep{Lu:DEFEAT:2023}} & \makecell{DFL\\SDFL} & Fully connected & \notick & \notick & \notick & \makecell{Secure aggregation} & \textit{N/S} & Simulated & Industry & \makecell{Federated models\\CPU usage} & \notick \\
\hline
\makecell{DFedSN\\\citep{Chen:DFedSN:2023}} & \makecell{DFL\\SDFL} & Fully connected & \maybetick & \yestick & \textit{N/S} & \makecell{Custom algorithm} & \makecell{TensorFlow\\Affine Transformations} & Simulated & Social Networks & \makecell{Federated models} & \notick \\
\hline
\makecell{\textbf{Fedstellar}\\(This work)} & \textbf{\makecell{DFL\\SDFL\\CFL}} & \textbf{\makecell{Fully connected\\Star, Ring,\\Random, Custom}} & \yestick & \yestick & \yestick & \makecell{Secure aggregation\\Communication encryption} & \textbf{\makecell{PyTorch\\Asynchronous sockets\\Docker containers}} & \textbf{Hybrid} & \textbf{Multiple} & \textbf{\makecell{Federated models\\Resources usage\\Communications}} & \yestick \\
\hline
\end{tabular}
\begin{tablenotes}
\item \yestick\space high,\space\maybetick\space moderate, \space\notick\space not addressed by the solution. \textit{N/S} (Not Specified) by the authors.
\end{tablenotes}
\end{threeparttable}}
\end{table*}

Starting from CFL architectures, TensorFlow Federated (TFF) served as the cornerstone in FL, offering a toolkit for federated computations \citep{google:tff:2019}. The TFF primarily aids server-based FL, a distinct characteristic of CFL, an approach highly beneficial for specific use cases such as smart grid systems adopting a star topology. In contrast, Federated AI Technology Enabler (FATE) successfully connected and integrated elements of CFL and distributed computation in centralized networks, a feature typically associated with DFL \citep{fedai:fate:2021}. However, FATE enables server-based FL by integrating communications optimization and secure computation. Finally, FedML emerged as a prominent FL platform oriented to academic purposes, offering support for various architectures and providing APIs for researchers \citep{He:fedml:2020}. It not only offers flexibility and scalability but also emphasizes maintaining network connections, a feature crucial for IoT devices where network access may occasionally be unreliable. However, none of these solutions fully support DFL, highlighting their struggle to adapt to diverse network configurations beyond the star topology.

After these CFL solutions, DFL marked its emergence with BrainTorrent, a pioneering solution in healthcare applications \citep{Roy:braintorrent:2019}. It functions in a Peer-to-Peer (P2P) setting, facilitating direct interaction between entities without a central coordinator. It is a valuable strategy when establishing a central trusted entity is challenging or a central server failure could halt the training process. BrainTorrent has proven successful, specifically for whole-brain segmentation of Magnetic Resonance Imaging (MRI) scans, presenting performance similar to traditional server-based FL and models trained on pooled data. However, the scope of BrainTorrent remains restricted, specifically tuned for healthcare applications, and it lacks the flexibility to adapt to various use cases and network topologies. Extending this P2P approach, the Scatterbrained framework expands the scope of DFL \citep{Wilt:scatterbrained:2021}. The framework separates the ML model from the FL business logic, reducing complicated tasks for developers and allowing the use of various learning tools. Its Python library simplifies the development process, providing an easy-to-use API to customize the parameter-sharing behavior. Although it provides substantial support for academic research, the primary focus on this area may restrict its practical applicability in real-world IoT environments. \addtxt{Another P2P-based alternative, p2pfl \citep{Guijas:p2pfl_report:2022}, enables customized topologies through DFL among training nodes. Upon setup, it randomly selects nodes for the whole federation using a votation. Despite lacking approaches like SDFL or CFL, it boasts significant extensibility and adaptability. Consequently, Fedstellar has incorporated aspects of its architecture into its federated models and node communications.}

In an increasingly privacy-conscious world, FL-SEC, a private and DFL architecture that excels in the Artificial Intelligence of Things (AIoT), has made significant advances \citep{Qu:fl-sec:2022}. It employs a unique blockchain structure to achieve decentralization, eliminates single points of failure, and protects against poisoning attacks. It also has a personalized incentive mechanism that encourages participation and reduces communications in many scenarios, like smart homes with fully connected topologies. Despite its advancements, the reliance of FL-SEC on blockchain technology may introduce complexity and scalability issues that Fedstellar mitigates with its scalable design. Further enhancing privacy is P4L, a privacy-preserving P2P learning system that uses multiple heterogeneous devices \citep{Arapakis:P4L:2023}. P4L utilizes partial Homomorphic Encryption (HE) to uphold the confidentiality of shared gradients, thereby circumventing information leakage attacks. Notably, the system operates without a centralized federation, Public Key Infrastructure (PKI), or internet connection, offering a cost-effective, privacy-preserving alternative to CFL architectures. Its performance has been evaluated through analytical simulations using CIFAR-10, Avito, and IMDB datasets under a fully connected topology, demonstrating that P4L can deliver federated models with competitive performance. However, it lacks the flexibility to incorporate new datasets easily and is restricted to a specific network topology. Furthermore, it does not provide sufficient metrics for monitoring the performance and progress of the learning process, which could be a limitation when used in more complex, real-world scenarios compared to Fedstellar.

Recognizing the need for security in smart industrial facilities, 2DF-IDS introduces a secure, decentralized, and Differentially Privacy (DP) FL-based Intrusion Detection System (IDS) \citep{Friha:2df-ids:2023}. It showcases its strength in smart industrial facilities utilizing random and ring topologies. It comprises three elements: a key exchange protocol, a DP gradient exchange scheme, and a decentralized FL approach. Its effectiveness in identifying various cyberattacks on an Industrial Internet of Things (IIoT) system using different datasets demonstrates its impressive capabilities. In this sense, 2DF-IDS customizes a gossip-based strategy to enhance its communication efficiency in decentralized environments \citep{Hashemi:gossip_comms:2022}. This strategy involves multiple gossip steps to navigate communication constraints, ensuring efficient gradient information exchange across the network. However, the solution has limitations, such as a lack of flexibility due to the limited customization of parameters such as aggregation algorithms, models, datasets, or network topologies. In addition, it does not facilitate actual deployments on physical devices. The DEFEAT framework, in its quest to defend against gradient attacks, uses a P2P network to transmit model parameters, enabling multiple connected clients to train a personalized model jointly \citep{Lu:DEFEAT:2023}. It balances communication costs and model training accuracy by using SDFL approaches with different schemes regarding message exchange. Evaluations on real datasets showed that DEFEAT performs better than other solutions in model training accuracy and effectively mitigates gradient attacks while maintaining high accuracy in the resulting models. As for the architecture of the solution, it does not present a modular structure and, therefore, lacks interdependence between the features of the solution. This limits the customization potential of the solution. In addition, the communications presented by the participants do not guarantee an asynchronous exchange of messages during federation, unlike Fedstellar, which keeps it implicit and transparent to the device.

Lastly, the DFedSN framework is an innovative solution adapted for different data types in social networks \citep{Chen:DFedSN:2023}. This framework adopts SDFL to establish data differences via affine transformation. It considers that all data users transmit as a unified image collection, fostering shared image classifications. This approach ensures that large data differences between users will not lead to performance problems during the federation. To address the challenges of communication efficiency, DFedSN adopts a balanced approach to minimize communication and computing costs, utilizing a decentralized approach where local computations are prioritized over frequent data exchanges \citep{Liu:balancing_compression_comms:2022}. Additionally, it incorporates an event-triggered method, enabling nodes to only transmit their local updates when significant changes in the model occur, which are determined by predefined thresholds \citep{Zehtabi:event_thresholds_comms:2022}. Tested on TensorFlow with two standard image datasets (MNIST and CIFAR-10) and implemented on three standard network architectures (AlexNet, Inception-Net, and Mini-ResNet) and a fully connected topology, DFedSN obtained results close to 95\% accuracy in the federated models, even with large differences in the distribution of user data.

In conclusion, while the solutions discussed herein contribute significantly to FL, they also present noteworthy limitations. These range from a lack of modular architecture and inadequate support for asynchronous message exchange during federation to limited customization and flexibility, such as a lack of support for different aggregation algorithms, models, datasets, or network topologies. Many of these DFL solutions are not open source, limiting their accessibility and adaptability. Furthermore, several solutions are restricted in scope to specific applications or fail to perform deployments on physical devices. Another significant limitation lies in the inability of many solutions to monitor real-time metrics, such as network traffic, which is crucial in decentralized scenarios. While these solutions have undoubtedly advanced the field, novel platforms are still needed to cover these important limitations.
\section{Fedstellar Platform}\label{sec:platform}

\change{Fedstellar is an innovative platform that}{Fedstellar, which extends from p2pfl library \citep{Guijas:p2pfl:2022},} facilitates the training of FL models in a decentralized fashion across many physical and virtualized devices \citep{MartinezBeltran:fedstellar_git:2023}. In particular, Fedstellar offers the following key features:

\begin{itemize}
\item Creation and management of federations comprising diverse types and numbers of real and/or virtualized devices\change{.}{, defining a simpler procedure than other solutions \citep{Roy:braintorrent:2019,Guijas:p2pfl:2022}.}
\item Generation and deployment of complex network topologies to link within a federation, including DFL, SDFL, and CFL, each catering to specific roles of the participants.\addtxt{ The base models of federation and inter-node communications have been extended from p2pfl library \citep{Guijas:p2pfl:2022} for compatibility with SDFL and CFL approaches. Also, Fedstellar implements leadership management, an improved workflow, the definition of roles, and the creation of network topologies on demand.}
\item Usage of different ML/DL models and datasets to tackle heterogeneous FL-based problems.
\item Incorporation of user-friendly features that enable flexible customization and efficient monitoring of training processes.
\item Monitoring of KPIs across multiple levels, including:
\begin{itemize}
\item \textit{Resource KPIs.} Monitoring of computational load, memory usage, and local model accuracy.
\item \textit{Communication KPIs.} Focus on communication latency, data transmission speed, and successful data transfer rate.
\item \textit{Federated model KPIs.} Assessment of model accuracy, precision, recall, and $F_{1} \ score$, training time, and convergence speed of each local model.
\end{itemize}
\end{itemize}

With these extensive features and integrated KPIs, Fedstellar effectively manages various learning scenarios and facilitates continuous performance assessment and optimization.

\subsection{Overall Architecture}

\begin{figure}[!htb]
\includegraphics[width=\columnwidth]{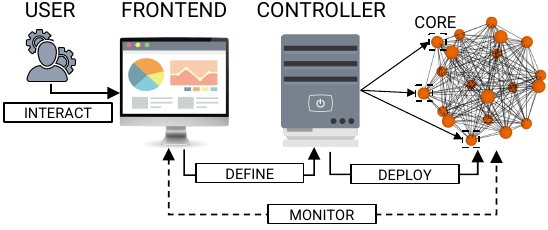}
\centering
\caption{Components of the Fedstellar platform: user, frontend, controller, and core}
\label{fig:overview}
\end{figure}

The architecture of Fedstellar unifies numerous functional components to facilitate the deployment of different FL architectures across physical and virtual devices through an intricate design encompassing advanced data handling, efficient training mechanisms, and communication protocols. The choice of Python as the underlying programming language is fundamental to the design, highlighting its capabilities in supporting diverse network topologies, managing efficient communication, and offering a rich ecosystem of ML and data processing libraries. As shown in Figure~\ref{fig:overview}, the architecture highlights four main components:

\begin{itemize}
\item \textit{User.} It has the responsibility of managing the platform operation. Through the intuitive interface of the frontend, users can establish, configure, and monitor federated scenarios, tailoring the system to the requirements of the learning process.
\item \textit{Frontend.} This interactive interface is where the user designs and oversees learning scenarios. Its user-focused design simplifies system configuration and supports ongoing KPIs performance tracking.
\item \textit{Controller.} The controller serves as the orchestration center of the platform. It interprets users' commands from the frontend, manages the entire federated scenario, assigns learning algorithms and datasets, and configures network topologies to ensure an efficient and effective FL process.
\item \textit{Core.} It is deployed on each device of the federation and is responsible for executing FL tasks\change{.}{, being partially extended from p2pfl library \citep{Guijas:p2pfl:2022}.} It manages model training, data preprocessing, secure communication among devices, and storage of the federated models. Additionally, the core supervises the calculation of KPIs and conveys this information back to the frontend for performance monitoring.
\end{itemize}

\begin{figure}[!htb]
\includegraphics[width=\columnwidth]{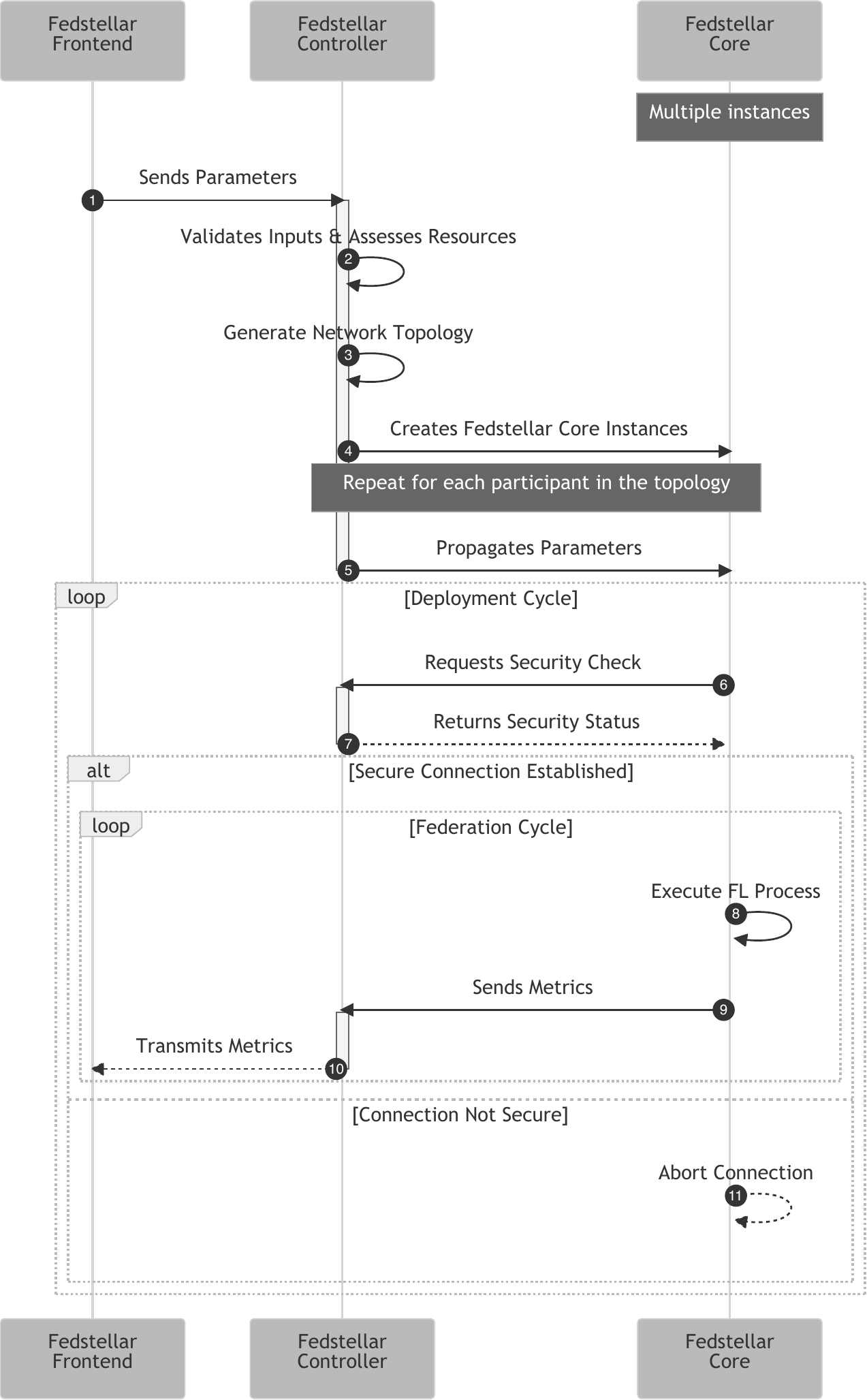}
\centering
\caption{Sequence diagram showing the interaction of Fedstellar components}
\label{fig:sequence-diagram}
\end{figure}

The various components of Fedstellar are meticulously intertwined, fostering a customizable and dynamic federated scenario. Also, multiple instances of the frontend and controller can be maintained simultaneously throughout the federation, ensuring continuous control and monitoring of deployed devices. This redundancy enhances the robustness of the system, minimizing downtime and maintaining operational integrity even during individual component failures. As shown in \figurename~\ref{fig:sequence-diagram}, the deployment process commences with the configuration of the scenario. Upon receiving the configuration from the frontend, the controller generates the network topology and establishes core instances for physical and virtualized devices. Then, it securely transfers the previously defined configuration to each device. This ensures a consistent and secure provision across all devices. The process culminates with the execution of the FL operation while monitoring and processing critical metrics that furnish insights into the performance and efficiency of the scenario.

\subsection{Fedstellar Frontend}
\label{sec:frontend}

\begin{figure*}[!htb]
\includegraphics[width=\textwidth]{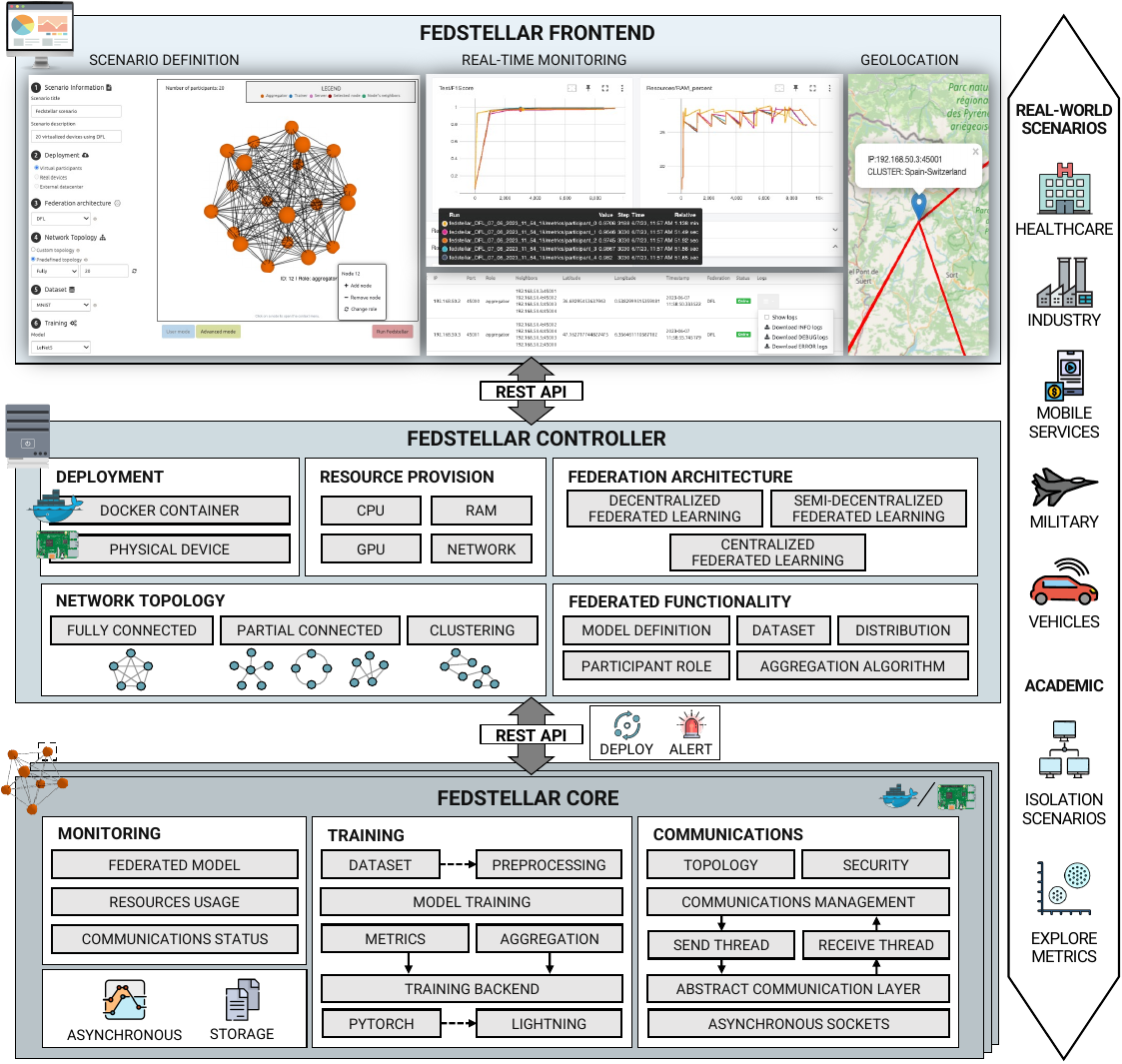}
\centering
\caption{Overall architecture of the Fedstellar platform}
\label{fig:implementation}
\end{figure*}

This component is designed to be user-friendly and intuitive, serving as the user's primary point of interaction with Fedstellar. As illustrated in the upper level of \figurename~\ref{fig:implementation}, users can define federated scenarios according to their needs, monitor their progress, and track the physical locations of devices within the topology. Additionally, a REST API ensures smooth communication with the controller, offering customizable endpoints for these functionalities.

\subsubsection{Scenario Definition}
\label{sec:scenario-definition}

A scenario within Fedstellar represents a unique execution of an FL experiment, distinguished by specific parameters, computational resource allocation, network topology, and targeted applications or devices. Defining a scenario involves setting parameters, extending beyond the basic naming and description for reference purposes. For instance, users can specify the deployment mode (physical or virtualized devices) and choose the federation architecture (DFL, SDFL, or CFL). Additionally, a key feature allows users to graphically design and interconnect nodes, which the system then converts into the corresponding adjacency matrix for managing device communication during the FL process. This feature is powered by D3.js, a JavaScript library known for its robust data-driven document manipulation and visualization capabilities \citep{Bostock:d3js:2021}.

In an advanced configuration view, users can define the specific dataset for the learning process, catering to various tasks, including image recognition, natural language processing, or predictive analytics. Users also have the flexibility to select the learning model, from traditional ML models to advanced DL models. Fedstellar employs a dynamic form creation system using a form generator based on JSON Schema to accommodate these diverse and intricate user inputs, ensuring adaptability and resilience across various FL scenarios.

\subsubsection{Real-time Monitoring}

The real-time monitoring functionality within Fedstellar plays a critical role, allowing users to track the performance and progress of ongoing federated scenarios. This feature equips users with up-to-the-minute updates on a range of metrics generated by each device within the network topology (see Section~\ref{sec:monitoring}). By offering this real-time insight, users can make timely, informed decisions, intervene when needed, and dynamically gauge the effectiveness of the experiment.

An important aspect of this capability is its customized implementation of TensorBoard, a suite extensively used for visualizing ML experiments \citep{google:tensorboard:2019}. In this sense, the suite has been optimized for handling simultaneous metric updates from numerous devices, ensuring rapid and accurate real-time data representation. Metrics loading time experiences a substantial reduction, significantly benefiting user experience in large-scale federations. Furthermore, the integration of a novel compression mechanism for TensorBoard events at the controller level leads to more efficient utilization of network resources and faster visualization rendering. Moreover, Fedstellar incorporates an extensible Logger functioning as an adapter to assure compatibility with various ML/DL libraries. This Logger transforms the metrics generated into a format compatible with TensorBoard, ensuring a seamless integration.

Additionally, the platform maintains a separation between the definition of metrics and the logging library in use. This design decision enhances the straightforward extension of the adapter to support other popular logging libraries, such as Wandb, MLFlow, or Neptune. Simultaneously, the platform ensures that data export capabilities remain robust, allowing users to extract the generated data in universally compatible formats such as CSV or JSON. This functionality enables seamless data integration with various data analysis and visualization tools, granting users the opportunity for more detailed exploration and understanding of the performance of their scenarios.

\subsubsection{Geolocation}

This functionality elevates the scope of experimentation and offers insightful context about the devices engaged in a federated scenario. This dynamic geolocation capability is further enhanced using Leaflet, an open-source JavaScript library known for its user-friendly interactive maps \citep{Agafonkin:leaflet:2021}. It enables the visualization of the distribution of the nodes and provides essential information such as each IP address and the cluster to which it belongs. This function maps out the precise geographical locations of physical devices, an essential feature in various analyses and applications where device positioning significantly impacts the outcome. This is particularly relevant for scenarios involving mobility or location-based factors, such as Unmanned Aerial Vehicle (UAV) fleets or vehicle-to-vehicle communication, in which the spatial distribution of devices is crucial for optimizing communication protocols and ensuring efficient data aggregation while preserving network integrity. Additionally, this feature aids in studying node mobility within a network, capturing data on devices as they enter or exit the federation, or tracking connectivity status during sporadic disconnections.

When utilizing virtual devices, users can manually assign specific geographical coordinates, enabling the simulation of virtual device geolocation. This capability plays a crucial role in facilitating dynamic emulation of node mobility, which is particularly relevant for applications dependent on factors such as node proximity or the ability to adapt to network topology changes swiftly. To ensure a comprehensive and realistic simulation, the platform leverages advanced network simulation libraries like tcconfig \citep{Hombashi:tcconfig:2023}. This integration allows the virtual geographical coordinates to influence system properties, including bandwidth and node latency. Consequently, users can conduct realistic evaluations in heterogeneous and dynamic environments. Moreover, users can track device geolocation using an interactive map on the frontend. With the added capability of a REST API, this feature ensures smooth integration with third-party applications and promotes effortless cross-platform geolocation monitoring.


\subsection{Fedstellar Controller}

As an integral component of Fedstellar, the controller serves as a bridge, enabling seamless interaction between the frontend and deploying core modules in physical or virtualized devices. Corresponding to the second level in \figurename~\ref{fig:implementation}, it includes the deployment and resource provision, network topology, federation architecture, and federated functionality, each playing a vital role in orchestrating the FL process. Furthermore, a REST API ensures robust connectivity between the controller and each deployed participant.

\subsubsection{Deployment and Resource Provision}

\begin{figure}[!htb]
\includegraphics[width=\columnwidth]{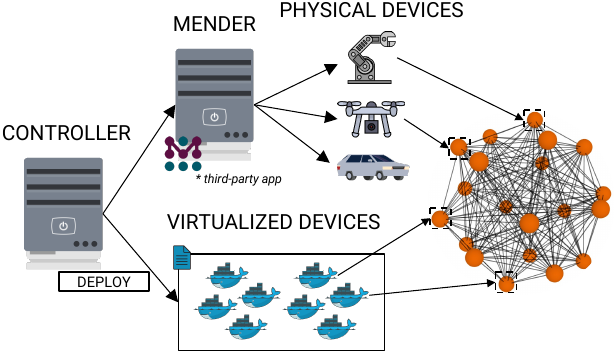}
\centering
\caption{Types of deployment of Fedstellar: virtualized and physical scenarios}
\label{fig:backend-deployment}
\end{figure}

This functionality adeptly blends state-of-the-art device management and virtualization technologies to accommodate both physical devices and virtual settings (see \figurename~\ref{fig:backend-deployment}). Beginning with physical devices, Fedstellar harnesses the powerful functionalities of Mender, a renowned open-source software purpose-built updater for embedded Linux devices \citep{Northern:mender:2021}. This strategic integration sets the way for a robust, efficient, and fail-safe software deployment process, smoothing the progression from a theoretical, experimental setup to a tangible FL application. Focusing on virtualized scenarios, Fedstellar taps into the capabilities of Docker containers. These containers are renowned for their innate potential for isolation, portability, and maintaining application consistency. They enable the encapsulation of applications and their dependencies into standardized units, ensuring a streamlined, replicable deployment process. In addition, Docker-level virtualization enables Fedstellar to deploy virtualized devices on a wide range of operating systems, including Windows, Linux, and MacOS.

Following the scenario definition, Fedstellar brings into play its resource provision management functionality. This feature is instrumental in the equitable distribution and regulation of computational resources among all participating nodes, be they physical or virtual devices. Participants receive a customizable YAML configuration file crafted to facilitate individual adjustments to their internal operations. This configuration encompasses a variety of settings, ranging from CPU allocation (including core count and percentage usage) and GPU settings (number of GPUs and usage strategies for simultaneous use) to RAM, disk, and network configurations (including subnet, IP address, and network conditions). Additionally, the configuration can also encompass settings related to federation, monitoring, and geolocation, all of which are user-defined via the frontend (refer to Section~\ref{sec:frontend} for more details). Regarding resource assignment and limitation, Fedstellar provides versatile and user-centric capabilities managed by REST API endpoints. For virtual devices, users can allocate available resources based on their specific requirements. In contrast, the allocation for physical devices depends on the inherent capabilities of the device, efficiently determined using Mender. To uphold these configurations, Fedstellar integrates a systemd-run mechanism to limit resources, fostering proficient and native resource management.

\subsubsection{Network Topology}

The creation of network topologies commences through a well-structured, user-oriented process. As commented in Section~\ref{sec:scenario-definition}, users define the initial topology within the intuitive frontend interface. In order to facilitate this process, the user can select preestablished topologies including, but not limited to, fully connected, partially connected (star, ring, random), or clustering configurations. Each selected topology corresponds to a specific configuration, and its selection activates a particular set of inter-device communication protocols. Once selected, Fedstellar creates a virtual internal network mirroring the user's selection, assigning distinct sockets to each participating device to enable seamless collaboration.

Fedstellar seamlessly manages dynamic changes in federated networks, easily integrating and adapting to nodes joining or leaving the federation. Each node connects to its neighbors through a network of sockets, and a continuous exchange of BEAT messages ensures uninterrupted monitoring of node availability. \addtxt{This type of message is included in \citep{Guijas:p2pfl:2022}, a well-known technique employed by other P2P solutions \citep{Wilt:scatterbrained:2021} as well.} A key feature of real-world FL scenarios is the unpredictable availability of devices, with nodes entering, exiting, or occasionally disconnecting from the network. The impact of these dynamic changes on training is contingent upon the topology and roles of the nodes. Critical nodes, such as proxies or aggregators, are pivotal because their exit can lead to communication disruptions or halt aggregation processes, respectively. Topologies like ring or star are especially vulnerable, with node exits potentially leading to device miscommunication \citep{MartinezBeltran:DFL_survey:2023}. However, the federated process is resilient, ensuring steady operations regardless of the network dynamics.

\subsubsection{Federation Architecture}

\begin{figure}[!htb]
\includegraphics[width=\columnwidth]{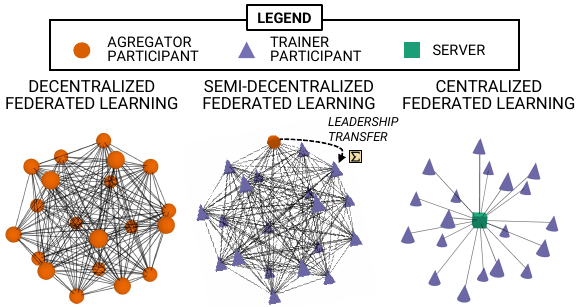}
\centering
\caption{Federation architectures implemented in Fedstellar: DFL, SDFL, and CFL}
\label{fig:federation-architectures}
\end{figure}

Concerning the federation architecture, users can select among DFL, SDFL, or CFL, as illustrated in \figurename~\ref{fig:federation-architectures}. DFL promotes a distributed environment wherein every device actively participates in local data training and aggregates the received model parameters from other devices. This collaborative approach bypasses the need for a central entity, minimizing the risk of a single point of failure and ensuring scalability. It is especially advantageous for expansive network structures where centralized control might pose challenges. SDFL is a hybrid approach that combines elements of both centralized and decentralized architectures. In this case, while most devices focus primarily on local training and parameter exchange, a single device is selected as the aggregator to combine model parameters from all participants. This dynamic aggregator role rotates among the devices at the end of each round or after a predefined time interval. Fedstellar implements a random selection mechanism but is open to integration with other schemes, depending on the scenario requirements. This approach balances centralized control and decentralized flexibility, ensuring a smoother flow in moderately complex network topologies. In contrast, CFL streamlines the process by employing a central server dedicated to model aggregation. This architecture provides efficiency and simplicity, ideally suiting less complex networks that can benefit from a centralized approach.

In terms of implementation, Fedstellar operates by associating a specific functionality to each participant based on the selected federation architecture during the deployment phase. In this sense, each participant is programmed with an independent thread for aggregating model parameters. Additionally, Fedstellar automatically generates leadership transfer messages using a custom-built Python class in SDFL. This class keeps track of the current leader and uses socket programming features to send leadership transfer messages to the next designated aggregator. In contrast, each participant is assigned to a unique set of thread lockers in CFL. These thread lockers, managed by the Python threading library, control the synchronization between the central server and participant devices, ensuring data integrity and orderliness in communications.

\subsubsection{Federated Functionality}

The setup process in Fedstellar commences with the user configuring various parameters from the frontend. This process entails selecting a suitable model structure from a comprehensive collection of ML/DL models, including architectures such as Deep Neural Network (DNN) and Convolutional Neural Network (CNN). These models are supported by well-established libraries such as PyTorch and Scikit-learn, granting users access to a broad array of algorithms. For users seeking a more tailored approach, Fedstellar also provides a means to integrate customized models, providing a template \addtxt{from \citep{Guijas:p2pfl:2022} }based on PyTorch Lightning and designed to ease integration. This template, built upon the foundation of the LightningModule, delineates inheritance functions for the definition of ML algorithms or DL layers, loss functions, optimizers, and performance metrics monitored during the learning phase.

Subsequently, the training dataset is defined. Fedstellar allows users to utilize standardized benchmark datasets, such as MNIST or CIFAR-10, and introduce their distinct datasets. For custom dataset incorporation, the platform provides a dedicated template designed to oversee the processing and management of the federation data effectively. This encompasses pivotal attributes such as dataset location (local or remote) and data preprocessing and augmentation considerations. Additionally, users can dictate the data distribution strategy, choosing between Independent and Identically Distributed (IID) or non-IID options. This choice has significant repercussions on partitioning the dataset among the devices. \addtxt{The platform extends the shuffle method proposed in \citep{Guijas:p2pfl:2022} with Dirichlet techniques and proportional class unbalancing. }To ensure effective data distribution, the platform first demarcates distinct portions of the dataset for each device by the selected distribution configuration. Subsequently, devices are directed to the dataset location, enabling them to load their designated portion based on pre-defined indices. While this data distribution functionality is a powerful tool for research and testing, simulating various data scenarios, Fedstellar is equally capable of supporting real-world deployments. Users aiming for practical applications must define their data acquisition method, ensuring versatility in catering to research needs and real-world scenarios.

Once these configurations are in place, Fedstellar initiates its federated functions. It recognizes three primary roles in the federated context: trainers, aggregators, and proxies. Trainers are responsible for local model training, aggregators for merging model updates, and proxies for collecting updates from trainers, alleviating the communication load on aggregators. Furthermore, users can select from an array of aggregation algorithms. The options span from Federated Averaging (FedAvg) to more specialized alternatives such as Krum, Trimmed Mean, or FedProx.

\subsection{Fedstellar Core}
This core is a fundamental component of the Fedstellar platform, \addtxt{being extended from \citep{Guijas:p2pfl:2022} and }deployed by the controller component into physical or virtualized devices (see the third level in \figurename~\ref{fig:implementation}). This element is responsible for managing most of the computation and communication required for the successful operation of FL. It is subdivided into several components: (i) the monitoring module for establishing the operational environment, (ii) the training module for handling the learning process, and (iii) the communications module for facilitating interactions between federated participants and with the frontend.

\subsubsection{Monitoring}
\label{sec:monitoring}

The monitoring capability of Fedstellar embodies a crucial component, facilitating systematic and periodic data collection about the ongoing federated scenario. This data collection process is executed every five seconds by default but can be customized to meet specific experimental requirements. With the aid of auxiliary libraries such as \textit{psutil} and \textit{pynvml}, Fedstellar proficiently gathers system-level and GPU-related metrics, respectively. The platform gathers metrics categorized into three primary groups:

\begin{itemize}
    \item \textit{Federated model.} This category includes metrics related to the model size, learning progress (encompassing loss, accuracy, precision, recall, and $F_{1}$ score), model synchronization frequency, and the number of training rounds.
    \item \textit{Resources usage.} It comprises parameters such as CPU usage, GPU usage, memory utilization, network bytes, and component temperatures, providing insights into the computational and storage resources consumed throughout the scenario.
    \item \textit{Communications status.} This category captures metrics related to communication operations, such as status updates, latency, active connections, and data transmission rates, illuminating network performance during federation.
\end{itemize}

In addition to the aforementioned aspects, Fedstellar takes responsibility for creating and managing log files generated during the execution of the FL process. These logs are a vital asset for potential troubleshooting, offering detailed insights into the operational attributes of the system. Additionally, they support the replication of scenarios and validation of results, reinforcing the scientific credibility of the conducted scenarios.

\subsubsection{Training}

This module is responsible for data management and model training within FL scenarios. \addtxt{This feature has been extended from p2pfl library \citep{Guijas:p2pfl:2022}, an open-source framework that allows the creation of federated models in P2P networks. In this sense, Fedstellar employs model hierarchy and aggregation control provided by the previous solution, redesigning the continuous interaction between nodes during federation and introducing more complex datasets and models. Also, Fedstellar provides additional configurations, such as aggregation algorithms and compatibility with ML models using the Scikit-learn library. }It utilizes a comprehensively organized directory, housing files integral to executing the specific federated scenario. Each of these files plays a critical role in outlining the dataset parameters, determining whether the data source is local or remote, orchestrating data transformations, and dictating the operation of the data loader. Preprocessing techniques like normalization, feature extraction, and data augmentation are deployed to refine the data quality further. These procedures play a pivotal role in processing raw data into a quality resource ready for model training, also providing a template to accommodate future data processing techniques.

Regarding model training, Fedstellar fosters a highly adaptable and expandable environment that eases the definition and modification of \change{learning}{ML and complex DL} models. By leveraging Lightning with PyTorch as a training backend, the platform simplifies complex ML logistics, accommodating diverse learning requirements corresponding to various datasets \citep{Falcon:PyTorch-Lightning:2019}. In terms of aggregation algorithms, Fedstellar provides a comprehensive suite of solutions, as highlighted in \tablename~\ref{tab:aggregation_algorithms}. This table briefly describes each algorithm and its compatibility with ML/DL models. Specifically, the aggregation algorithms can be applied to ML models such as Logistic Regression and SVM. In the case of Logistic Regression, this means averaging the gradients, while for SVM, it involves merging support vectors, which are the key data points defining the decision boundary. For DL models, such as NNs or autoencoders, the aggregation process averages the weights and biases across all layers from the federated models\change{.}{, being the case of FedAvg \citep{Guijas:p2pfl:2022}.} \change{The platform is built based on extensibility principles, enabling}{Following the same principle of extensibility as \citep{Guijas:p2pfl:2022}, the platform enables} users to incorporate custom aggregation algorithms or those imported from third-party libraries. This adaptability is especially beneficial for researchers and developers experimenting with novel aggregation techniques, fostering innovation and exploration within the FL ecosystem.

\begin{table}[ht!]
\caption{Aggregation algorithms supported by Fedstellar}
\label{tab:aggregation_algorithms}
\resizebox{\columnwidth}{!}{
\centering
\begin{tabular}{p{3.5cm} p{9cm} p{2cm}} 
\hline
\textbf{Algorithm} & \textbf{Description} & \textbf{Compatibility (ML/DL)} \\ 
\hline
\multicolumn{2}{c}{\textit{Standard Aggregation Algorithms}}
\\ \hline

\makecell[l]{FedAvg\\\citep{McMahan:FL_google:2016}} & A widely-adopted DFL aggregation algorithm. Each node computes model updates and collaboratively derives a weighted average $\omega_{global} = \frac{1}{N} \sum_{k=1}^{N} \omega_k$ & ML/DL \\ \hline

\makecell[l]{Median\\\citep{Pillutla:median_robust_agg:2022}} & Determines the median value of the model parameters across nodes. Effective in countering certain adversarial data manipulations. & ML/DL \\

\midrule
\multicolumn{2}{c}{\textit{Secure Aggregation Algorithms}}
\\ \midrule

\makecell[l]{Krum\\\citep{Blanchard:krum_agg:2017}} & Resists adversarial nodes by selecting the update with the smallest cumulative distance to other updates $f(N, B) = N - B - 2$, where $B$ denotes adversarial nodes among $N$ total nodes. & DL \\ \hline

\makecell[l]{Trimmed Mean\\\citep{Yin:trimmed_mean_agg:2018}} & Reduces the impact of extreme updates. Excludes the highest and lowest values, averaging the middle values $TM(x) = \frac{1}{n-2r} \sum_{i=r+1}^{n-r} x_i$ & ML/DL \\ \hline

\makecell[l]{FedProx\\\citep{Li:fedprox_agg:2020}} & Addresses challenges from nodes with non-IID data or those that have sporadic contributions by introducing a regularization term $L(\omega) + \mu |\omega - \omega_0|^2$ & DL \\ \hline

\makecell[l]{Zeno\\\citep{Xie:zeno_agg:2019}} & Removes potentially misleading updates, converging when the consensus among the valid updates is achieved. & DL \\ \hline

\makecell[l]{Fed+\\\citep{Kundu:fedplus_agg:2022}} & Evaluates the variance in local updates and adjusts the aggregation process to optimize outcomes. & DL \\

\hline
\end{tabular}}
\end{table}

\subsubsection{Communications}

\begin{figure}[!htb]
\includegraphics[width=\columnwidth]{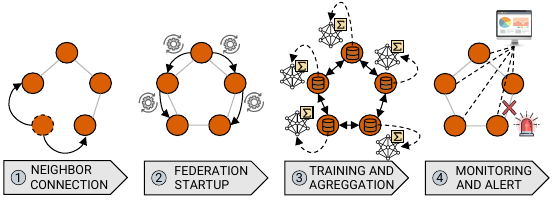}
\centering
\caption{Schematic representation of the communication process in Fedstellar}
\label{fig:steps}
\end{figure}

This module primarily communicates and transmits necessary information within the network, including the frontend for status updates and the neighboring nodes for the asynchronous exchange of model parameters. \addtxt{This last point has been improved from the version provided by p2pfl library \citep{Guijas:p2pfl:2022}, which enables the use of P2P communications to exchange information between peers. Fedstellar introduces improvements in compatibility and extensibility in virtual networks provided by Docker containers in virtual scenarios and with Mender in the deployment of physical scenarios. In addition, transparent network usage management and geolocation are included with fault-tolerant connections, allowing a reconnection or link change in case of congestion or node location. }\tablename~\ref{tab:comms_messages} outlines the messages deployed in Fedstellar, providing a detailed account of the communication process. \addtxt{Most of them are obtained from \citep{Guijas:p2pfl:2022}.} Following this exposition, as depicted in \figurename~\ref{fig:steps}, the communication process has four steps: (i) connecting all nodes with their neighbors; (ii) initiating the federation setup; (iii) executing the training process with decentralized aggregation; and (iv) monitoring and alerting.

\begin{table}[ht!]
\caption{Communication messages implemented in Fedstellar}
\label{tab:comms_messages}
\resizebox{\columnwidth}{!}{
\centering
\begin{tabular}{l l} 
\hline
\textbf{Message} & \textbf{Description} \\ 
\hline
\multicolumn{2}{c}{\textit{Messages with forwarding}}
\\ \hline

START\_LEARNING & Initiates the FL process \\
STOP\_LEARNING & Halts the ongoing FL process \\
PARAMS & Exchange of model parameters among participants \\
STOP & Signals termination of the current federation scenario \\
MODELS\_READY & Indicates readiness of locally trained models \\
MODELS\_AGGREGATED & Informs about the completion of the aggregation \\

\hline

\multicolumn{2}{c}{\textit{Messages without forwarding}}
\\ \hline
CONNECT\_TO & Requests a node to establish a connection with another \\
BEAT & Validates the connection and liveliness of a node \\
ROLE & Assigns a role (trainer, aggregator, or proxy) to a node \\
METRICS & Transfers collected monitoring data for analysis \\
LEADERSHIP & Transfers the aggregation functionality in SDFL \\
\hline
\end{tabular}}
\end{table}

During the initial step, each device within the federation creates a communication link with its immediate neighbors, storing these connections through open sockets. This step ensures the efficient and prompt exchange of information throughout the federation. Significantly, these communications are executed atop an abstract layer, offering the versatility to employ other technologies for data exchange. Moreover, Fedstellar incorporates advanced encryption mechanisms for both the exchange of keys and the encryption of communicated messages. By leveraging a combination of symmetric and asymmetric encryption offered by the pycryptodome library, Fedstellar assures secure model exchanges alongside efficient key management \addtxt{provided by \citep{Guijas:p2pfl:2022} in P2P networks.} This strategic implementation ensures data confidentiality and delivers strong protection against possible breaches. In the second step, the federation initiation commences with a designated node broadcasting a message outlining the definition, including the federation rounds, which refer to the cyclical process of local model training and model update sharing within the federation. On receiving this message, nodes tailor their operations to align with their assigned roles and responsibilities within the federation. While the current design involves all available nodes, future enhancements could fine-tune node inclusion based on available resources or network characteristics, including latency and delay.

The third step of the process is the key to the FL operation: model training, decentralized aggregation, and asynchronous exchange of model parameters. In this step, each node employs its respective data to train local models independently. An implementation leveraging a thread-based system ensures the transparent, asynchronous receipt, forwarding, and sending of messages between nodes. This system employs incoming and outgoing message queues to maintain a structured flow of communication, storing and processing incoming and outgoing messages in an orderly manner. To enhance communication efficiency, the platform incorporates an adaptive gossip technique to forward messages through participants, dynamically adjusting the flow of transmissions based on network conditions and node availability. It facilitates the decentralized aggregation approach whereby the trained model parameters are asynchronously exchanged among neighboring nodes using reduced latency and bandwidth. The platform integrates an intuitive mechanism to signal when a device should perform aggregation. This action is triggered when all neighboring nodes have transmitted their parameters or after a default timeout period of five minutes. Additionally, the platform offers a user-friendly functionality accessible through the frontend, enabling precise customization of the timeout period to accommodate the unique characteristics and requirements of each device in the federation. This ensures a smooth and timely workflow. The types of messages employed in this process are outlined in \tablename~\ref{tab:comms_messages}.

The concluding step of the process concerns monitoring and alerting. The federated participants actively assess and report on any disruptions, such as link failures or devices leaving the topology. In this way, nodes can remove their affected neighboring nodes from their internal queue of open sockets. Furthermore, the nodes maintain unidirectional connections with the frontend through secure WebSockets, facilitating the continuous transmission of metric data and ensuring constant federation monitoring. Moreover, Algorithm~\ref{alg:cycle} provides a comprehensive exposition of the internal cycle of each participant, aligning with the steps mentioned above.

\begin{algorithm}[ht!]
\caption{Federated participant cycle in Fedstellar}
\label{alg:cycle}
\footnotesize
\begin{algorithmic}[1]
\Require{$R$: local round, $\alpha$: learning rate, $\lambda$: regularization parameter, $S_j$: socket to neighbor j, $D$: local dataset}

\State $D_{\text{Train}}, D_{\text{Test}} \gets split(D)$

\For{$r$ in $R$}

    \State $\text{Initialize Local Model with Parameters } \theta$ 
    \For{each $(x, y)$ in $D_{\text{Train}}$}
        \State $\theta \gets \theta - \alpha (\nabla_\theta J(\theta, x, y) + \lambda \theta)$ \Comment{\textbf{Train}}
    \EndFor
    
    \For{$j$ in $N$} \Comment{\textbf{Send}}
        \State $\text{Model Parameters to } j \text{ via } S_j$ 
    \EndFor 
    
    \While{$not \; Timeout$}
        \For{$j$ in $N$} \Comment{\textbf{Receive}}
                \State $RP_j \gets \text{Model Parameters from } j
                \text{ via } S_j$ 
        \EndFor
    \EndWhile
    
    \State $\theta \gets \frac{1}{|N|+1} (\theta + \sum_{j \in N} RP_j)$ \Comment{\textbf{Aggregate} (FedAvg)}
    \State $\text{Update Local Model with } \theta$

\EndFor

\For{each $(x, y)$ in $D_{\text{\textbf{Test}}}$}
    \State $y_{pred} \gets \text{Predict with Local Model on } x$ \Comment{\textbf{Test}}
    \State $L \gets \frac{1}{|D_{\text{Test}}|}\sum_{i=1}^{|D_{\text{Test}}|} l(y, y_{pred})$ \Comment{\textbf{Compute Loss}}
\EndFor

\State $\text{Send metrics to frontend}$ \Comment{\textbf{Report Metrics}}

\end{algorithmic}
\end{algorithm}

\section{Validation Scenarios}
\label{sec:validation}

The Fedstellar platform has been deployed on two federated scenarios: a physical scenario, which showcases the real-world applicability of Fedstellar, highlighting its performance in a real scenario with tangible resource-constrained devices such as Raspberry Pis; and a virtualized scenario, which illustrates the scalability of the platform and its ability to handle a wide variety of configurations. \tablename~\ref{tab:validation-scenarios} compares the characteristics of both scenarios.

\begin{table}[ht]
\caption{Validation scenarios of Fedstellar using physical and virtualized devices}
\label{tab:validation-scenarios}
\resizebox{\columnwidth}{!}{
\centering
\begin{tabular}{c c c}
\hline
\textbf{Characteristic} & \textbf{Physical Scenario} & \textbf{Virtualized Scenario} \\ 
\hline
\makecell{Participant\\Characteristics} & 8 (5 Raspberry Pi 4 / 3 Rock64) & 20 (docker containers) \\
\hline
Dataset & Syscalls \citep{Huertas:syscall-dataset:2023} & \makecell{MNIST \citep{Deng:MNIST:2012} /\\CIFAR-10 \citep{krizhevsky:cifar10-100:2009}} \\
\hline
\makecell{Federated\\Model} & Autoencoder & LeNet5 / MobileNet \\
\hline
\makecell{Network\\Topology} & \makecell{Fully connected,\\Star, Ring} & \makecell{Fully connected,\\Star, Ring} \\
\hline
\makecell{Federation\\Architecture} & DFL & DFL, SDFL, CFL \\
\hline
\end{tabular}}
\end{table}

\subsection{Overview}

The physical scenario leveraged the IoT spectrum sensors from ElectroSense, an open-source, real-world crowdsensing platform \citep{Rajendran:electrosense:2018}. These low-cost sensors were employed to collect data from the radio frequency spectrum. Each sensor was calibrated to monitor ``normal'' behavior. Further manipulations of the source code led to the execution of eight types of Spectrum Sensing Data Falsification (SSDF) attacks per sensor: Delay, Confusion, Freeze, Hop, Mimic, Noise, Repeat, and Spoof \citep{Huertas:ssdf-attacks:2023}. These attacks were designed to alter the spectral data observed by the sensors, allowing for the detection and identification of malicious behavior.

In contrast, the virtualized scenario was focused on image recognition. Given the immense volume of unlabeled image data ubiquitously present on devices, this serves as a significant scenario for deploying FL capabilities. The widely-used MNIST and CIFAR-10 datasets served as resources for this scenario, enabling the validation of the Fedstellar platform in a frequently explored academic and industry context. This emphasizes the application of FL in enhancing the accuracy and efficiency of image classification tasks, a significant step forward in developing AI technologies.

\subsection{Participant Characteristics}

In the physical scenario, the federation comprised eight devices, five of which were Raspberry Pi 4 boards, and the remaining three were Rock64 boards, all in the same private local network. The Raspberry Pi 4 devices, equipped with a 1.5GHz quad-core 64-bit ARM Cortex-A72 CPU and 2GB of RAM, presented an optimal combination of size, affordability, and processing power. Alternatively, the Rock64 boards enhanced the heterogeneity by offering a slightly inferior processing capacity, characterized by a 64-bit ARM Cortex-A53 with a 1.5 GHz clock speed and up to 2GB RAM.

In contrast, the virtualized scenario constituted twenty virtual devices as Docker containers. These containers were deployed on a host machine with an Intel Core i7-10700F processor that operates at a base frequency of 2.9 GHz (up to 4.8 GHz), 128GB RAM, 2TB storage, and dual NVIDIA RTX 3080 GPUs. The allocation of GPUs to virtual participants was achieved through a random selection scheme, where each GPU served ten participants. 

\subsection{Dataset and Federated Model}

In the physical scenario, a dataset containing syscalls monitored under SSDF attacks was employed \citep{Huertas:syscall-dataset:2023}. Syscalls are a potent source of data for detecting various types of cyberattacks due to their ability to capture the system internal actions and process anomalies \citep{Sanchez:fingerprinting:2021}. The dataset contains syscalls from the spectrum collection service and its processes over 60-second intervals. The federated model employed was an autoencoder configured with a DNN architecture and trained via the FedAvg aggregation algorithm. The autoencoder, comprised of 64, 16, 8, 16, and 64 neurons across its hidden layers, used the ReLU activation function for non-linear transformation. The model learning defined an anomaly recognition threshold set at the $95^{th}$ percentile of the reconstruction error of the training dataset. Thus, any deviation beyond this point is flagged as a potential threat.

For the virtualized scenario, Fedstellar was validated against the MNIST and CIFAR-10 datasets, both being well-established for benchmarking. The MNIST dataset, a robust collection of handwritten digits, is frequently used for training and validating models within the ML domain \citep{Deng:MNIST:2012}. Similarly, the CIFAR-10 dataset comprises images distributed across ten distinct classes with a more complex classification task \citep{krizhevsky:cifar10-100:2009}. To address the specific needs of these datasets, Fedstellar employed two different models: a LeNet5 model for the MNIST dataset \citep{Lecun:lenet5:1998}, and a MobileNet model for the CIFAR-10 dataset \citep{Howard:mobilenet:2017}, both lightweight yet powerful models. LeNet5, a pioneering Convolutional Neural Network (CNN), consists of several layers that recognize image patterns. MobileNet is a lightweight model for mobile and embedded vision applications. The aggregation of the federated models was executed using the FedAvg algorithm, similar to the physical scenario. Notably, the data used in physical and virtualized scenarios were non-IID, reflecting real-world conditions where data are often unevenly distributed across devices. In the experiments, participants underwent ten federation rounds, following the process detailed in Algorithm~\ref{alg:cycle}, with local models training for twenty epochs in each round. The ten federation rounds encompass the entire federated learning cycle, including parameter transmission and aggregation, while the twenty epochs pertain solely to the local training on each device.

\subsection{Network Topology}

In the physical scenario, a diverse range of network topologies were employed to validate the robustness and adaptability of Fedstellar, including fully connected, star, and ring configurations. The fully connected topology, fostering complete intercommunication among devices, ensures optimal information flow, which is crucial for effective DFL. The star topology demonstrates the versatility of the platform, as it introduces a centralized node for efficient information routing, suitable for scenarios where specific devices play a central role in the federation. Additionally, the ring topology highlights scalability and efficiency, minimizing communication overhead in larger federations.

The virtualized scenario leveraged the same network topologies as the physical scenario: fully connected, star, and ring. The implementation of these topologies in a virtual environment made it possible to simulate diverse federated learning contexts and network configurations, testing the adaptability and performance of the platform under various simulated conditions.

\subsection{Federation Architecture}

The DFL architecture was selected for the physical scenario due to its resilience and robustness. The DFL approach facilitates decentralization, eliminating potential bottlenecks and vulnerabilities tied to a central server. Its adoption offers valuable insights for future real-world implementations, particularly where intermittent connectivity and device failures are prevalent.

Conversely, the virtual scenario exploited the benefits of all three federation architectures: DFL, SDFL, and CFL. The variety was chosen to test the proficiency of Fedstellar under various conditions. SDFL allows Fedstellar to demonstrate its ability to balance decentralization and strategic centralization for optimal learning outcomes. The inclusion of CFL, on the other hand, underscores the ability of the platform to handle scenarios that demand a fully centralized structure, emphasizing its versatility.

\section{Results}\label{sec:results}

The evaluation of Fedstellar performance, encompassing both physical and virtualized deployments, focused on key indicators such as the $F_{1} \ score$ for federated models, the percentage of CPU and RAM usage, network traffic in megabytes $(\text{MB})$, and model convergence time. These measurements provide comprehensive insights across varying federated architectures, topologies, datasets, and ML/DL models.

\subsection{Physical Scenario}

\begin{figure*}[htb!]
  \centering
  \begin{subfigure}{.25\textwidth}
    \centering
    \includegraphics[width=\linewidth]{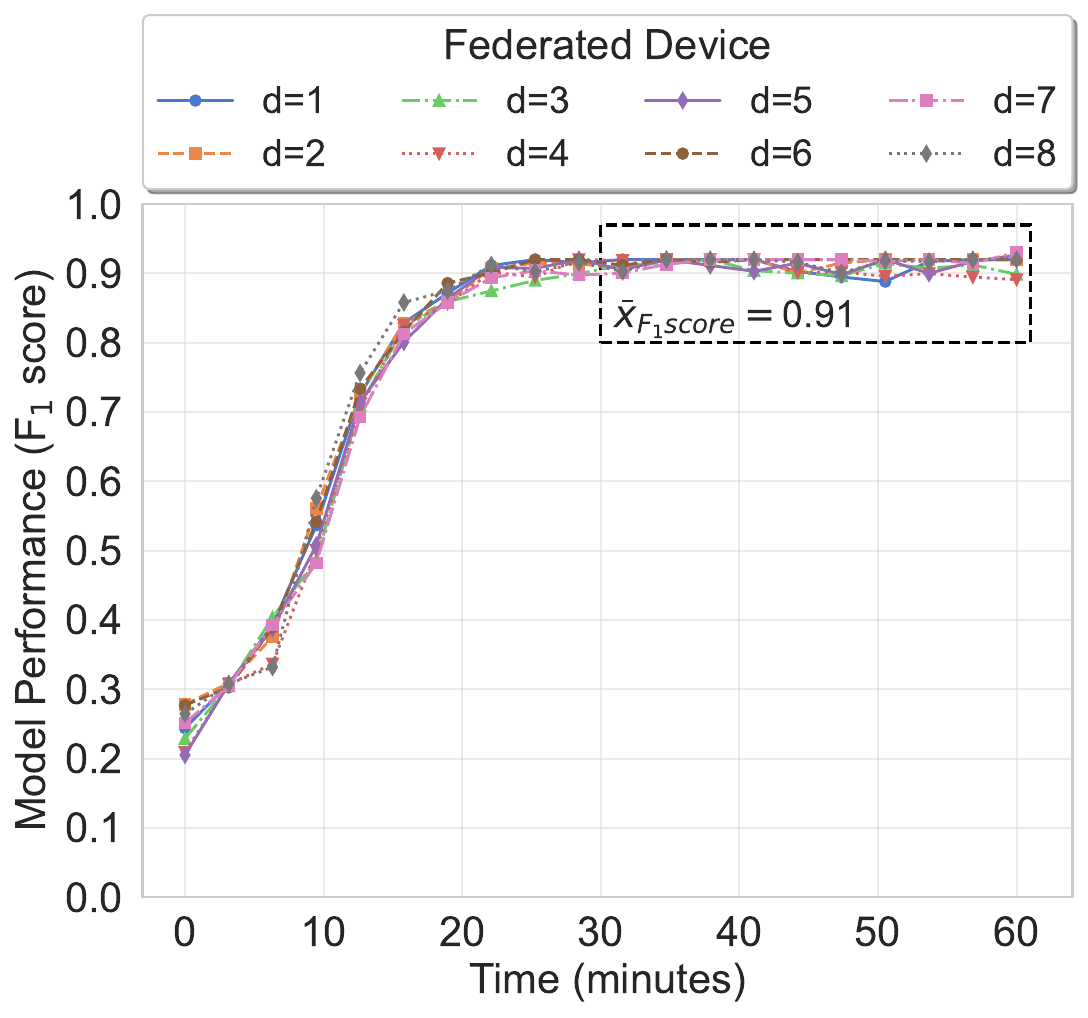}
    \caption{Federated models $(F_{1} \ score)$}
    \label{fig:results_physical_f1_fully}
  \end{subfigure}%
  \begin{subfigure}{.25\textwidth}
    \centering
    \includegraphics[width=\linewidth]{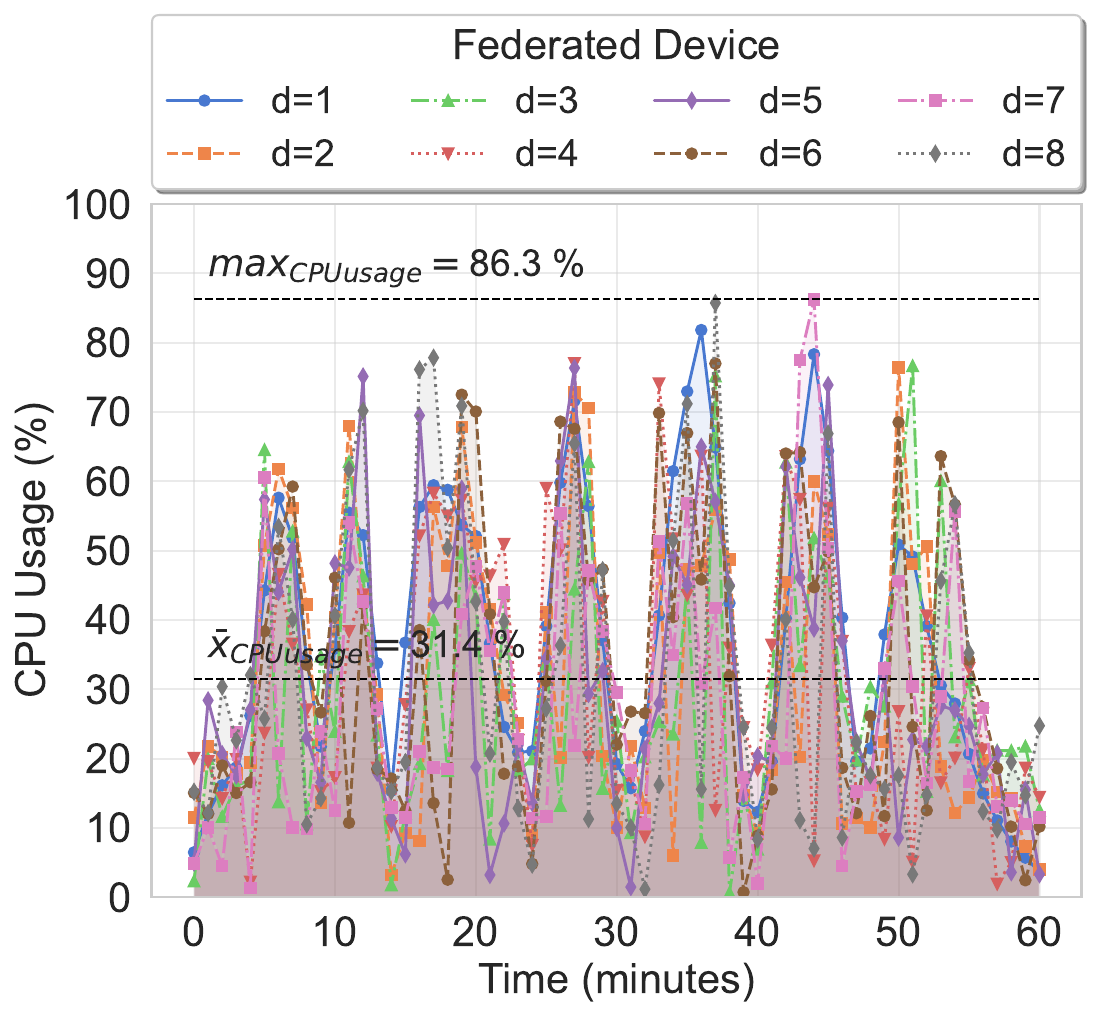}
    \caption{CPU usage $(\%)$}
    \label{fig:results_physical_cpu_fully}
  \end{subfigure}%
  \begin{subfigure}{.25\textwidth}
    \centering
    \includegraphics[width=\linewidth]{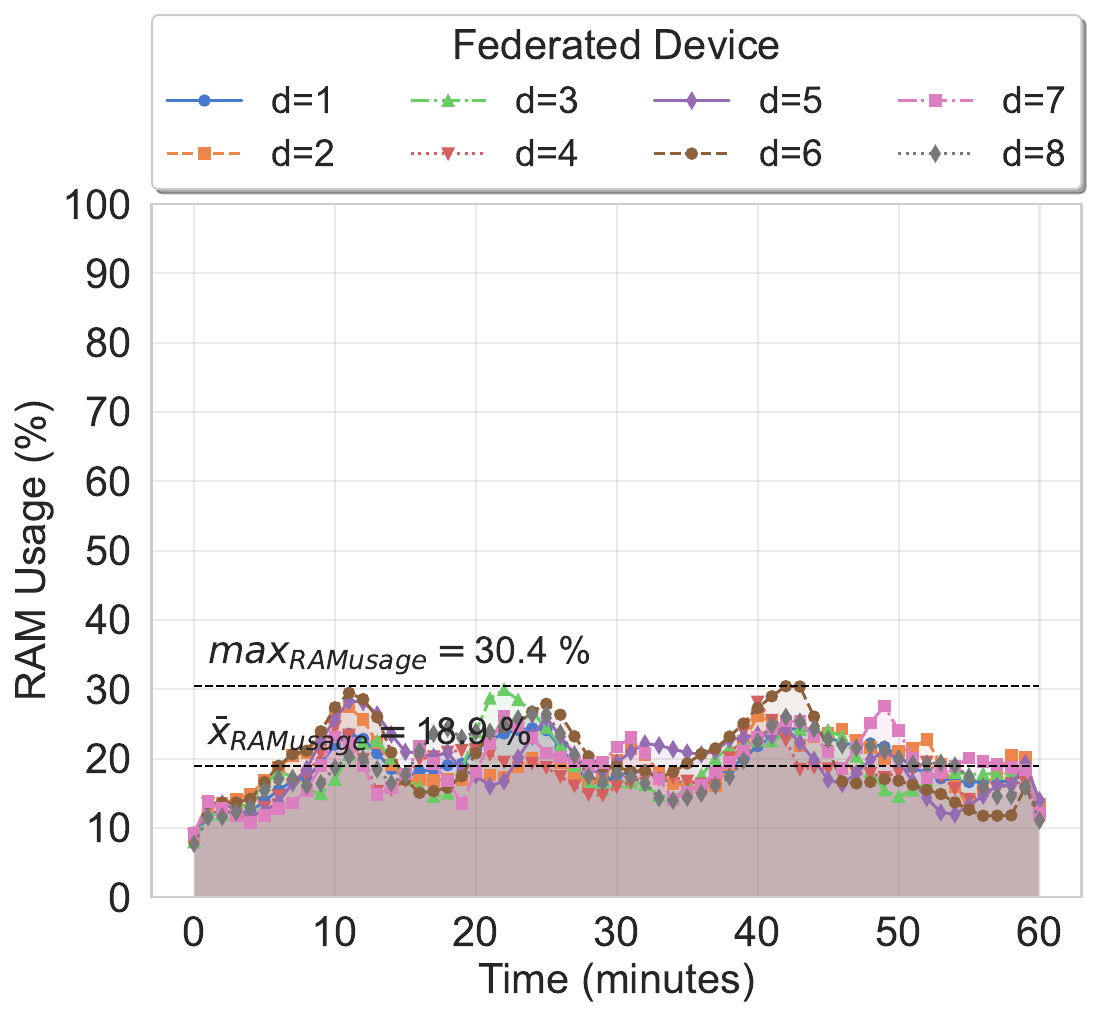}
    \caption{RAM usage $(\%)$}
    \label{fig:results_physical_ram_fully}
  \end{subfigure}%
  \begin{subfigure}{.255\textwidth}
    \centering
    \includegraphics[width=\linewidth]{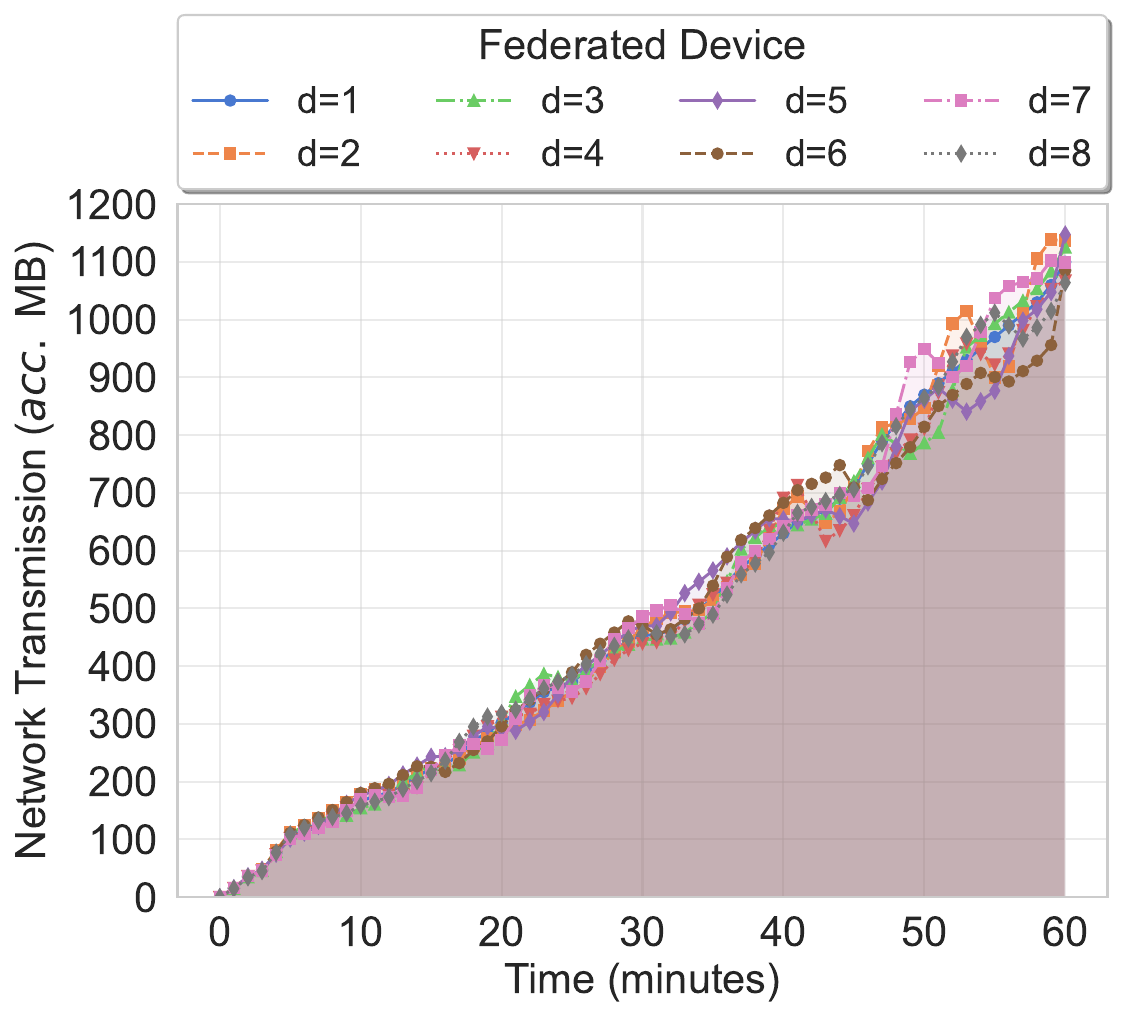}
    \caption{Network usage $(\text{MB})$}
    \label{fig:results_physical_net_fully}
  \end{subfigure}

  \caption{Performance of Fedstellar in a physical deployment utilizing a fully connected topology across eight devices, with data gathered over 60 minutes using a syscall dataset}
  \label{fig:results_physical_fully}
\end{figure*}

\begin{figure*}[htb!]
  \centering
  \begin{subfigure}{.25\textwidth}
    \centering
    \includegraphics[width=\linewidth]{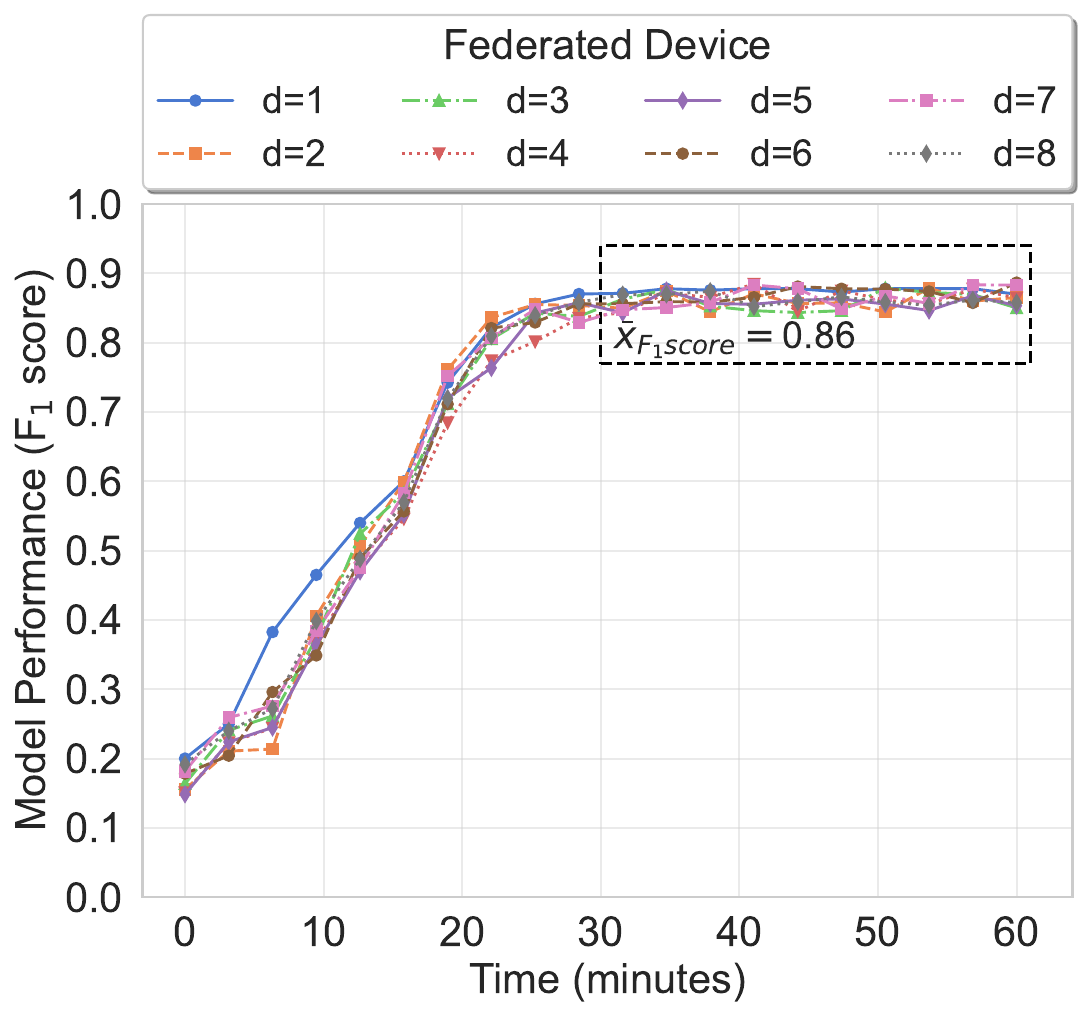}
    \caption{Federated models $(F_{1} \ score)$}
    \label{fig:results_physical_f1_star}
  \end{subfigure}%
  \begin{subfigure}{.25\textwidth}
    \centering
    \includegraphics[width=\linewidth]{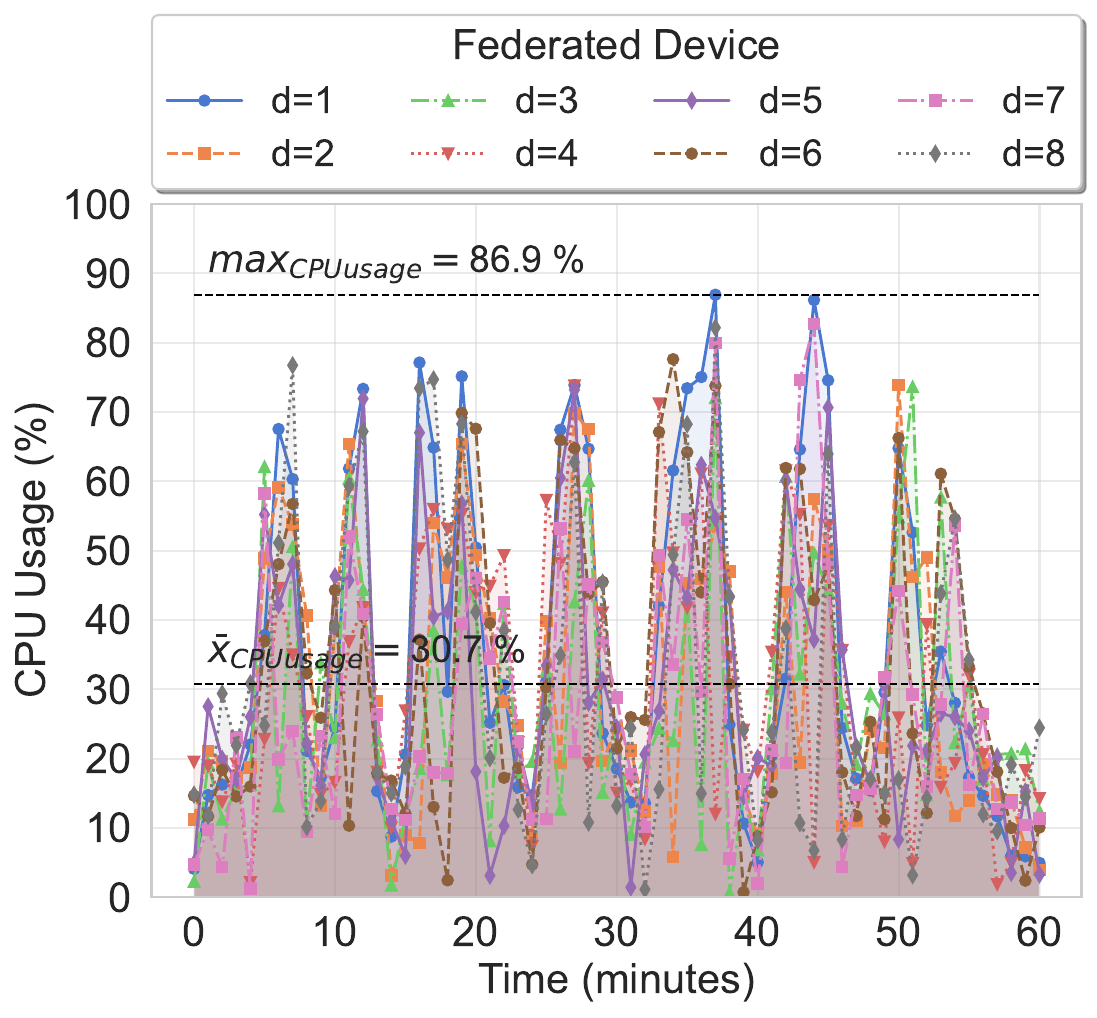}
    \caption{CPU usage $(\%)$}
    \label{fig:results_physical_cpu_star}
  \end{subfigure}%
  \begin{subfigure}{.25\textwidth}
    \centering
    \includegraphics[width=\linewidth]{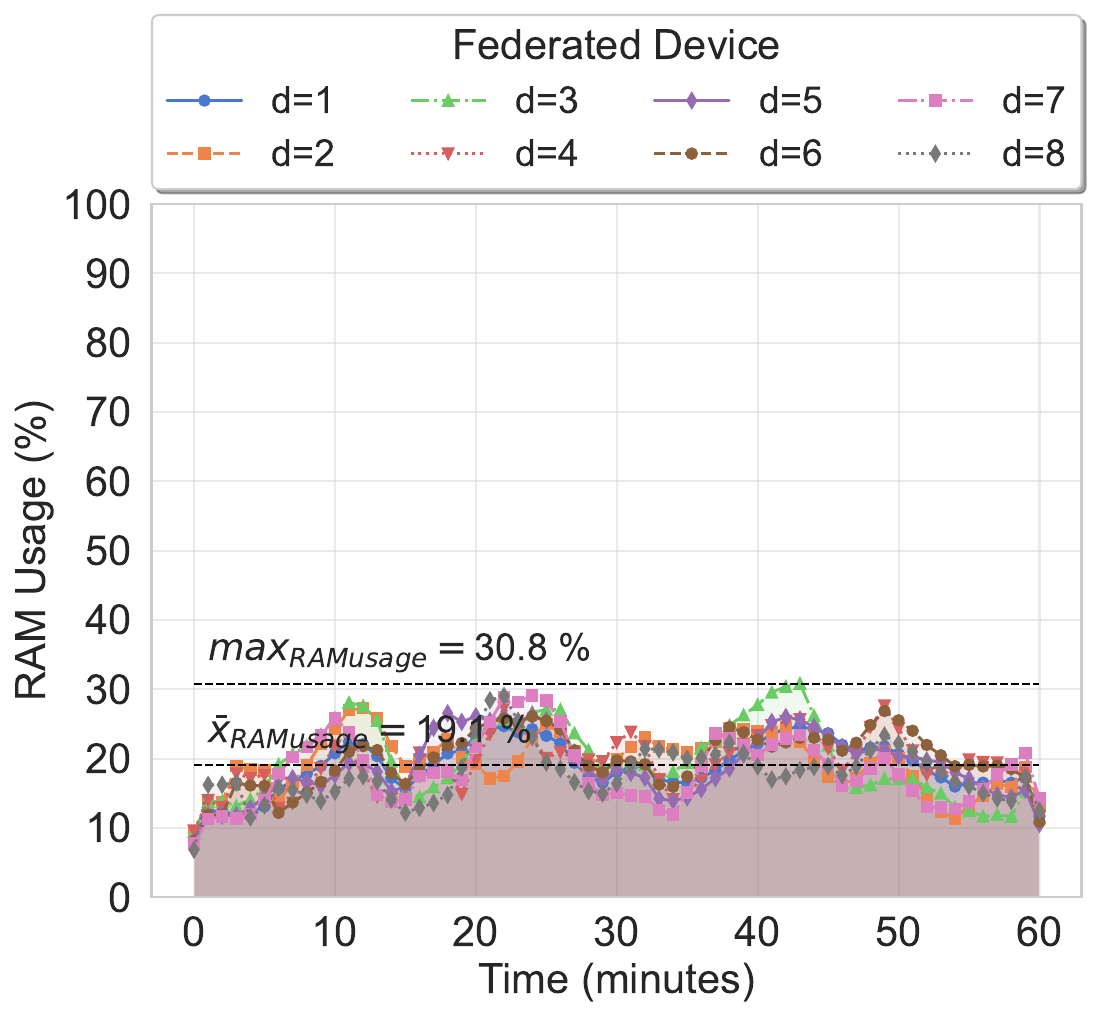}
    \caption{RAM usage $(\%)$}
    \label{fig:results_physical_ram_star}
  \end{subfigure}%
  \begin{subfigure}{.255\textwidth}
    \centering
    \includegraphics[width=\linewidth]{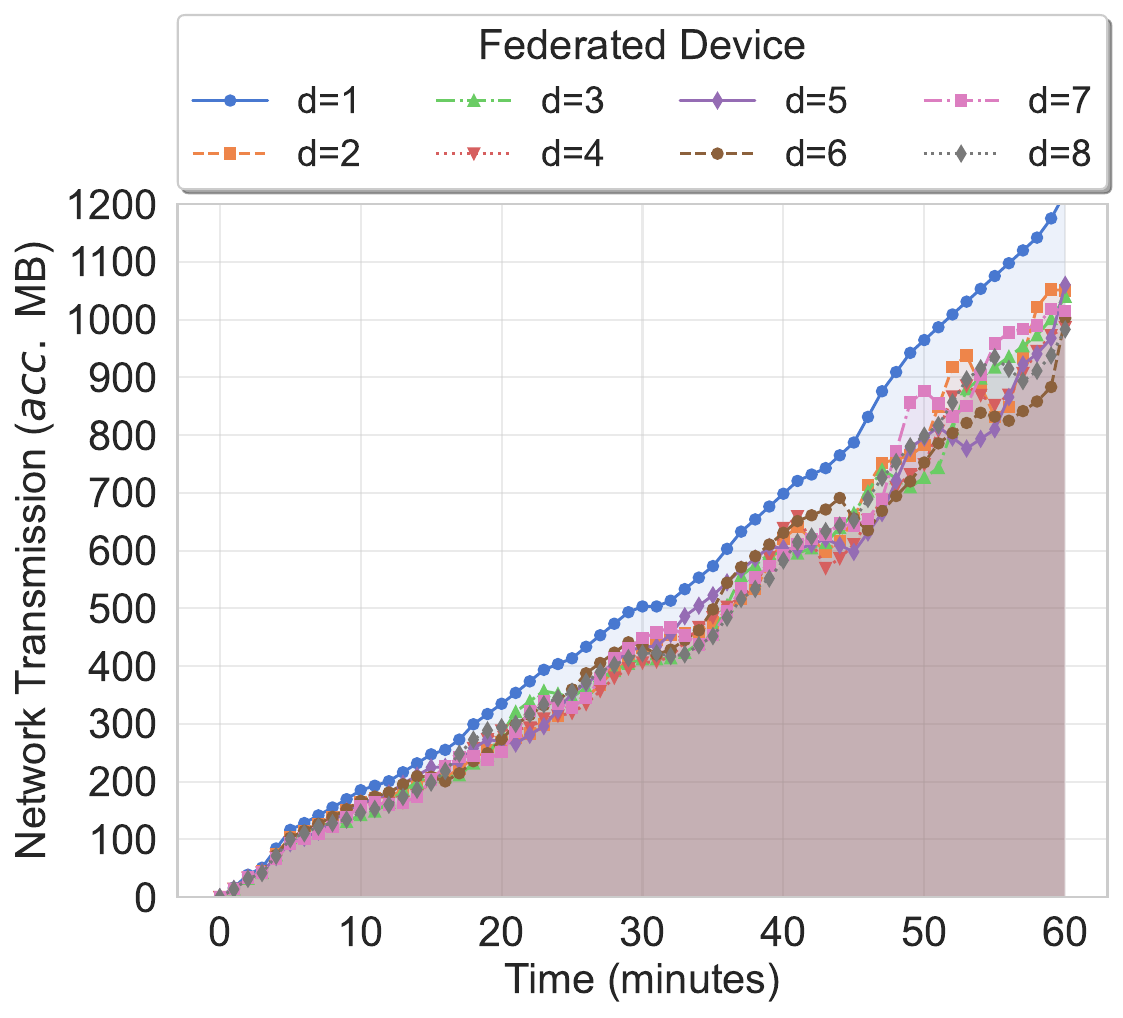}
    \caption{Network usage $(\text{MB})$}
    \label{fig:results_physical_net_star}
  \end{subfigure}

  \caption{Performance of Fedstellar in a physical deployment utilizing a star topology across eight devices, with data gathered over 60 minutes using a syscall dataset}
  \label{fig:results_physical_star}
\end{figure*}

\begin{figure*}[htb!]
  \centering
  \begin{subfigure}{.25\textwidth}
    \centering
    \includegraphics[width=\linewidth]{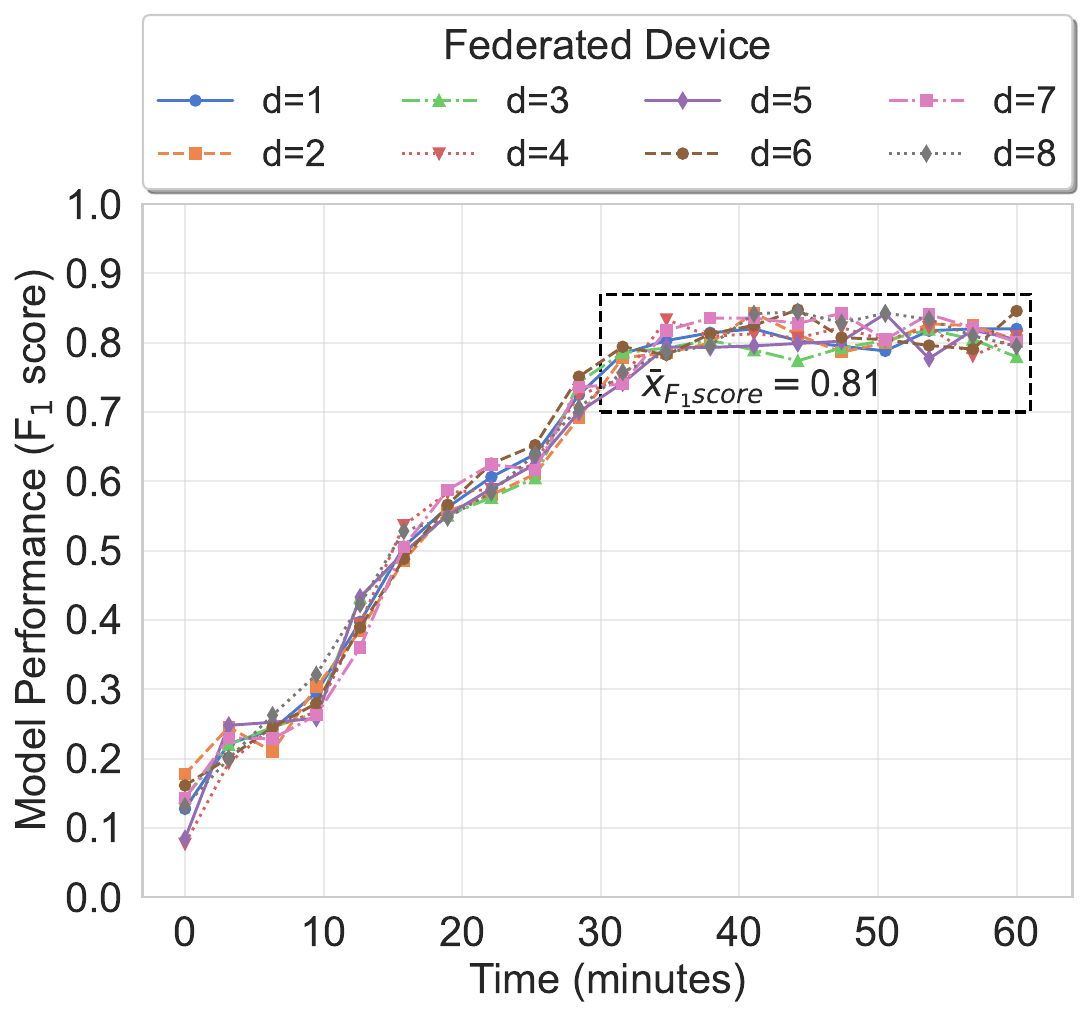}
    \caption{Federated models $(F_{1} \ score)$}
    \label{fig:results_physical_f1_ring}
  \end{subfigure}%
  \begin{subfigure}{.25\textwidth}
    \centering
    \includegraphics[width=\linewidth]{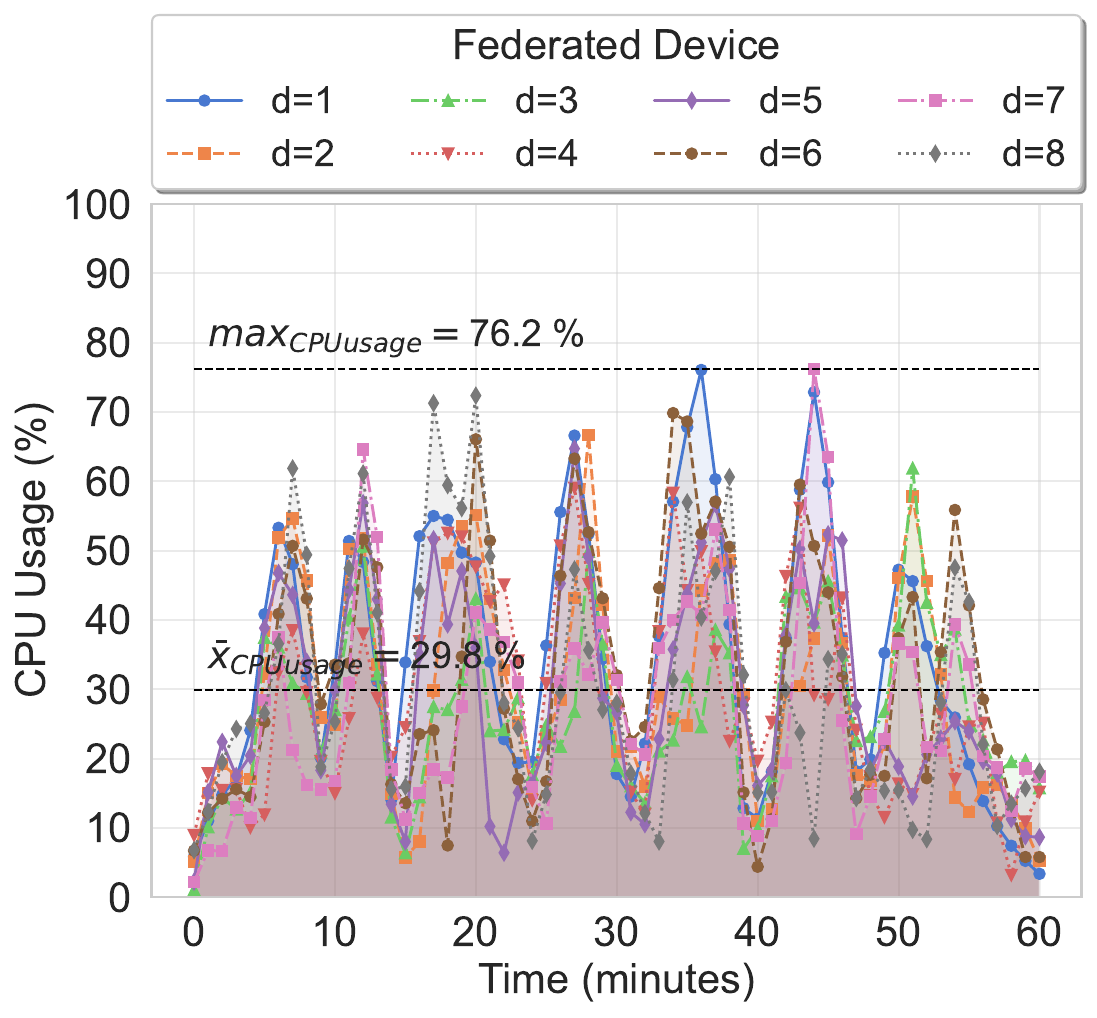}
    \caption{CPU usage $(\%)$}
    \label{fig:results_physical_cpu_ring}
  \end{subfigure}%
  \begin{subfigure}{.25\textwidth}
    \centering
    \includegraphics[width=\linewidth]{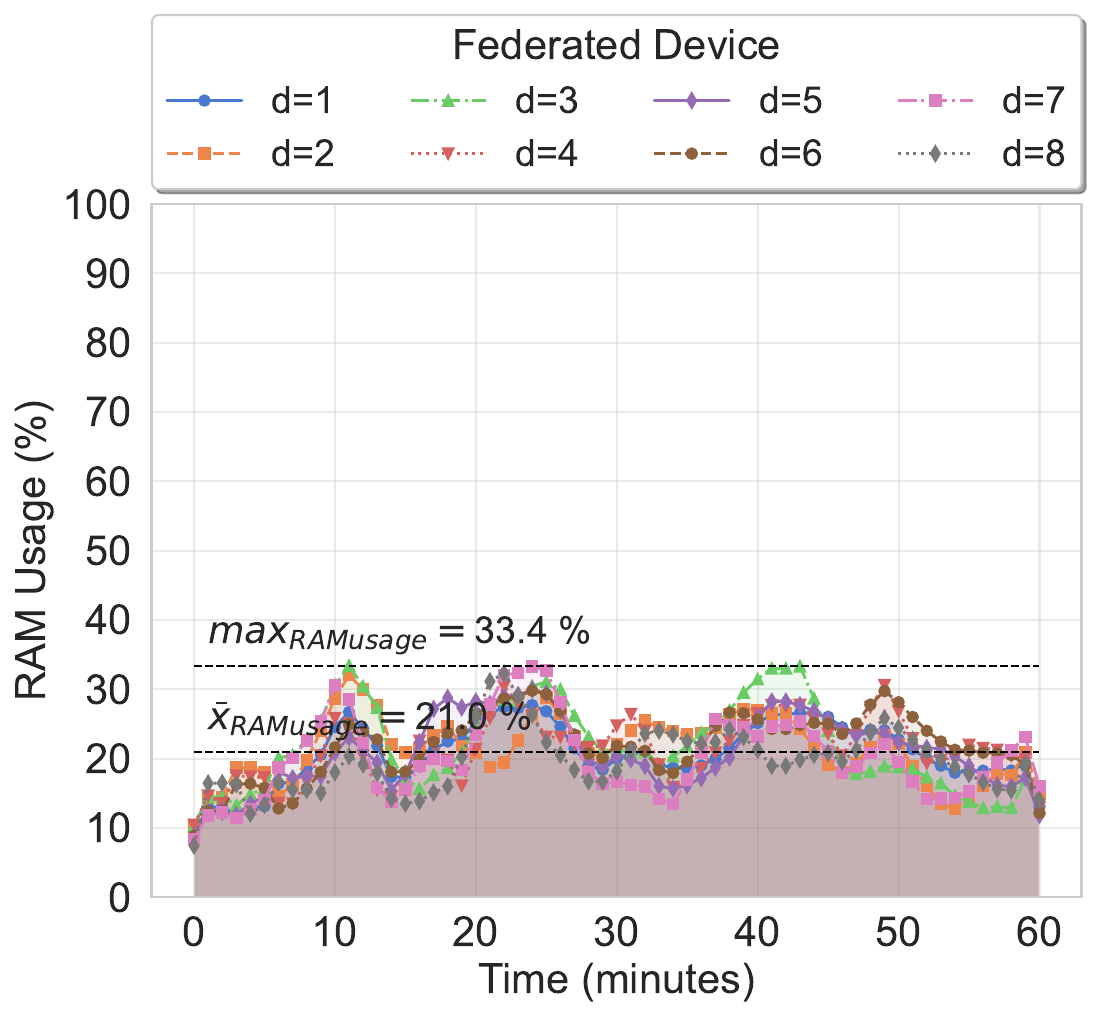}
    \caption{RAM usage $(\%)$}
    \label{fig:results_physical_ram_ring}
  \end{subfigure}%
  \begin{subfigure}{.255\textwidth}
    \centering
    \includegraphics[width=\linewidth]{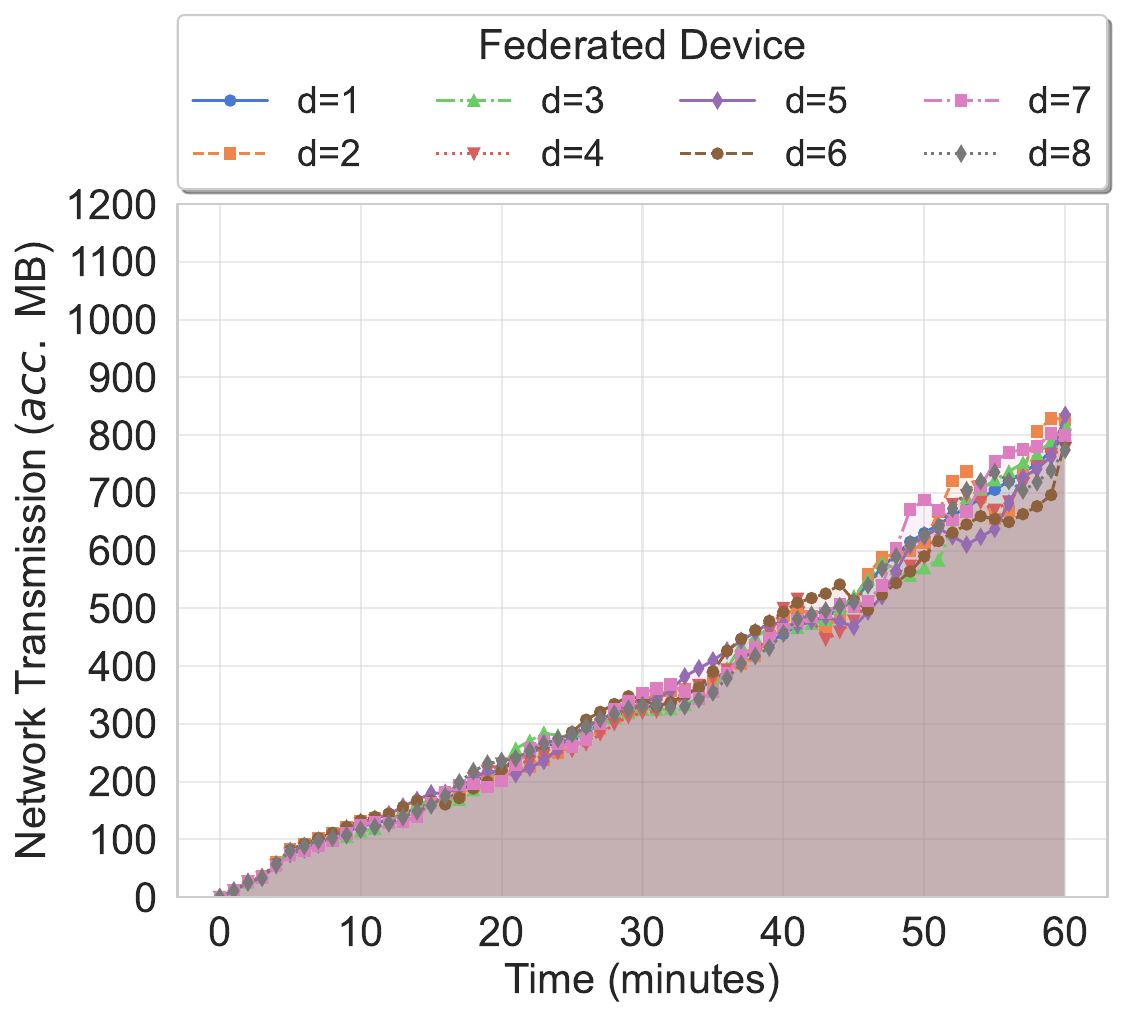}
    \caption{Network usage $(\text{MB})$}
    \label{fig:results_physical_net_ring}
  \end{subfigure}

  \caption{Performance of Fedstellar in a physical deployment utilizing a ring topology across eight devices, with data gathered over 60 minutes using a syscall dataset}
  \label{fig:results_physical_ring}
\end{figure*}

Delving into the details of the physical deployment, \figurename~\ref{fig:results_physical_fully}, \figurename~\ref{fig:results_physical_star}, and \figurename~\ref{fig:results_physical_ring} illustrate the outcomes of Fedstellar running on fully connected, star, and ring topologies with eight devices, each labeled as $d=\text{\{1-8\}}$. These devices, encompassing five Raspberry Pi 4 units (1-5) and three Rock64 units (6-8), operate in an intricately orchestrated network, showcasing the full capability of Fedstellar in real-world scenarios.

Regarding model performance, results present a compelling narrative of the learning efficiency within the network topologies. Within the first 15 minutes of training, the fully connected topology achieves an $F_{1} \ score$ of 80\% (see \figurename~\ref{fig:results_physical_f1_fully}). In comparison, the star topology takes slightly longer than 20 minutes (see \figurename~\ref{fig:results_physical_f1_star}), while the ring topology requires up to 30 minutes to reach the same benchmark (see \figurename~\ref{fig:results_physical_f1_ring}). Beyond these initial phases, the $F_{1} \ score$ in the fully connected topology continues its upward trajectory, peaking at 91\% by the end of an hour. Meanwhile, the star and ring topologies register 86\% and 81\%, respectively. Even though the eight devices used for the experiment have different hardware capabilities, they demonstrate similar performance patterns.

However, the asynchronous operations of Fedstellar introduced variability in the $F_{1} \ score$ among devices during the federation process. Devices in sparser topologies (see \figurenames~\ref{fig:results_physical_f1_star} and \ref{fig:results_physical_f1_ring}) experienced more fluctuations in performance due to less frequent parameter exchanges between neighbors. In contrast, fully connected topologies lead to a more uniform improvement across devices (see \figurename~\ref{fig:results_physical_f1_fully}). This asynchronization also resulted in subtle variances by the 60-minute mark, with some devices displaying slightly lower $F_{1} \ scores$ than others. The timeout mechanism in each device, designed to prevent delays from slower devices, contributes to this variability as higher throughput devices could trigger timeouts and perform local aggregations earlier. Furthermore, the random initialization of local models introduces an initial layer of variability, although it decreases as the models converge over time.

Moreover, \figurenames~\ref{fig:results_physical_cpu_fully}, \ref{fig:results_physical_cpu_star}, and \ref{fig:results_physical_cpu_ring} reveal similar patterns of CPU usage, specifically reflecting the highest utilized core across the eight devices involved. The patterns oscillate consistently, with cycles of CPU usage increases lasting 5-10 minutes, during which the utilization reaches a mean of 80\%, with an occasional peak of up to 86.9\% in the fully connected topology. Each of these surges is then followed by a drop, reducing the CPU usage to roughly 10-30\%. The correlation of these trends with the computational cycle outlined in \figurename~\ref{fig:steps} and Algorithm~\ref{alg:cycle} is remarkable. More in detail, intense CPU utilization coincides with periods of model training, and a notable reduction happens as the trained models undergo parameter exchanges with neighboring nodes.

Moving into RAM usage, results demonstrated a more irregular pattern. The inconsistency likely stems from the variation in data volume derived from syscalls in the system and, thus, the dataset creation. On average, there is a 20\% RAM usage in the three topologies, with sporadic spikes that hover between 30.4\%, 30.8\%, and 33.4\%. It is plausible to relate the RAM usage to the data handling demands of the DFL process, where each device processes its local dataset, adding to the memory footprint. The results highlight an upward trend in accumulated network usage across all topologies. In the fully connected topology, as depicted in \figurename~\ref{fig:results_physical_net_fully}, there is a significant rise in data exchange among devices, reaching a substantial consumption of 1190 MB by the close of the 60-minute duration. Comparatively, the star and ring topologies register 1050 MB and 830 MB, respectively, as illustrated in \figurename~\ref{fig:results_physical_net_star} and \figurename~\ref{fig:results_physical_net_ring}. Notably, in the star topology, the central node, designated as d=1, consistently records a higher data exchange throughput, culminating in 1200 MB at the endpoint. This progressive surge in network activity is primarily attributed to the continuous exchange of model parameters among nodes. As these interactions intensify, it results in a marked escalation in network traffic.

In conclusion, these results collectively validate the performance of the Fedstellar platform, demonstrating its capability to navigate the complexities of a decentralized IoT environment. It showcases a successful adaptation to asynchronous processes, efficient resource management, and the adept handling of extensive network activity. This suggests that Fedstellar is well-positioned to facilitate effective FL solutions in real-world IoT scenarios.

\subsection{Virtualized Scenario}
\label{sec:results-virtual}

In this scenario, Fedstellar illustrates its proficiency in adapting to varying configurations. With the deployment of twenty docker containers, the system handles different federation architectures and network topologies, each tested with the MNIST and CIFAR-10 datasets. The performance analysis, as shown in \tablename~\ref{tab:metrics-MNIST} and \tablename~\ref{tab:metrics-CIFAR10}, offers a detailed view of the average metrics obtained across the twenty containers over 60 minutes.

\begin{table}[ht!]
\caption{Average metrics obtained by twenty virtualized participants using MNIST dataset during 60 minutes}
\label{tab:metrics-MNIST}
\centering
\resizebox{\columnwidth}{!}{
\begin{threeparttable}
\begin{tabular}{c c c c c c c} 
\hline
\textbf{\makecell{Federation\\Architecture}} & \textbf{\makecell{Network\\Topology}} & \makecell[r]{\textbf{Model}\\$(F_{1} \ score)$} & \makecell[r]{\textbf{CPU}\\$(\%)$} & \makecell[r]{\textbf{RAM}\\$(\%)$} & \makecell[r]{\textbf{Network}\\$(\text{MB})$} & \makecell[r]{\textbf{Time*}\\$(min.)$} \\
\hline
\multirow{3}{*}{DFL} & Fully & \makecell{0.987 $\pm0.009$} & \makecell{78 $\pm15\ \%$} & \makecell{29 $\pm6 \ \%$} & \makecell{$\approx 1243 \ \text{MB}$} & \makecell{$\approx 28$} \\
& Star & \makecell{0.955 $\pm0.012$} & \makecell{72 $\pm13\ \%$} & \makecell{28 $\pm5\ \%$} & \makecell{$\approx 1165 \ \text{MB}$} & \makecell{$\approx 35$} \\
& Ring & \makecell{0.917 $\pm0.019$} & \makecell{70 $\pm14\ \%$} & \makecell{26 $\pm4\ \%$} & \makecell{$\approx 1089 \ \text{MB}$} & \makecell{$\approx 41$} \\
\hline
\multirow{3}{*}{SDFL} & Fully & \makecell{0.973 $\pm0.015$} & \makecell{69 $\pm12\ \%$} & \makecell{28 $\pm5\ \%$} & \makecell{$\approx 1148 \ \text{MB}$} & \makecell{$\approx 32$} \\
& Star & \makecell{0.938 $\pm0.020$} & \makecell{66 $\pm11\ \%$} & \makecell{27 $\pm4\ \%$} & \makecell{$\approx 1065 \ \text{MB}$} & \makecell{$\approx 38$} \\
& Ring & \makecell{0.901 $\pm0.027$} & \makecell{64 $\pm13\ \%$} & \makecell{25 $\pm4\ \%$} & \makecell{$\approx 1023 \ \text{MB}$} & \makecell{$\approx 45$} \\
\hline
CFL & Star & \makecell{0.992 $\pm0.010$} & \makecell{58 $\pm10\ \%$} & \makecell{26 $\pm3\ \%$} & \makecell{$\approx 985 \ \text{MB}$} & \makecell{$\approx 40$} \\
\hline
\end{tabular}
\begin{tablenotes}
\item * Overall time to reach model $F_{1} \ score \geq 90\ \%$
\end{tablenotes}
\end{threeparttable}}
\end{table}

\begin{table}[ht!]
\caption{Average metrics obtained by twenty virtualized participants using CIFAR-10 dataset during 60 minutes}
\label{tab:metrics-CIFAR10}
\centering
\resizebox{\columnwidth}{!}{
\begin{threeparttable}
\begin{tabular}{c c c c c c c} 
\hline 
\textbf{\makecell{Federation\\Architecture}} & \textbf{\makecell{Network\\Topology}} & \makecell[r]{\textbf{Model}\\$(F_{1} \ score)$} & \makecell[r]{\textbf{CPU}\\$(\%)$} & \makecell[r]{\textbf{RAM}\\$(\%)$} & \makecell[r]{\textbf{Network}\\$(\text{MB})$} & \makecell[r]{\textbf{Time*}\\$(min.)$} \\
\hline
\multirow{3}{*}{DFL} & Fully & \makecell{0.912 $\pm0.012$} & \makecell{80 $\pm16\ \%$} & \makecell{32 $\pm6\ \%$} & \makecell{$\approx 1280 \ \text{MB}$} & \makecell{$\approx 33$} \\
& Star & \makecell{0.889 $\pm0.015$} & \makecell{76 $\pm14\ \%$} & \makecell{30 $\pm6\ \%$} & \makecell{$\approx 1202 \ \text{MB}$} & \makecell{$\approx 38$} \\
& Ring & \makecell{0.862 $\pm0.020$} & \makecell{74 $\pm15\ \%$} & \makecell{28 $\pm5\ \%$} & \makecell{$\approx 1134 \ \text{MB}$} & \makecell{$\approx 43$} \\
\hline
\multirow{3}{*}{SDFL} & Fully & \makecell{0.895 $\pm0.016$} & \makecell{72 $\pm13\ \%$} & \makecell{29 $\pm6\ \%$} & \makecell{$\approx 1156 \ \text{MB}$} & \makecell{$\approx 36$} \\
& Star & \makecell{0.873 $\pm0.020$} & \makecell{70 $\pm14\ \%$} & \makecell{28 $\pm5\ \%$} & \makecell{$\approx 1090 \ \text{MB}$} & \makecell{$\approx 40$} \\
& Ring & \makecell{0.852 $\pm0.027$} & \makecell{67 $\pm12\ \%$} & \makecell{27 $\pm6\ \%$} & \makecell{$\approx 1050 \ \text{MB}$} & \makecell{$\approx 46$} \\
\hline
CFL & Star & \makecell{0.924 $\pm0.013$} & \makecell{62 $\pm11\ \%$} & \makecell{28 $\pm4\ \%$} & \makecell{$\approx 1020 \ \text{MB}$} & \makecell{$\approx 43$} \\
\hline
\end{tabular}
\begin{tablenotes}
\item * Overall time to reach model $F_{1} \ score \geq 85\ \%$
\end{tablenotes}
\end{threeparttable}}
\end{table}

\tablename~\ref{tab:metrics-MNIST} details the performance with the MNIST dataset, where the CFL configuration reports the highest $F_{1} \ score$ of 99.2 \%. Although this high score emphasizes the efficiency of a centralized approach, it is essential to acknowledge the trade-offs involved. Despite reporting a slightly lower $F_{1} \ score$ of 98.7 \%, the DFL with a fully connected topology presents considerable advantages. A key highlight is its accelerated convergence, taking approximately 28 minutes to reach an $F_{1} \ score$ of 80 \%, a considerable reduction in time compared to the 40 minutes of CFL. This acceleration in achieving model performance indicates an increased efficiency integral to the DFL platform despite a slight trade-off in $F_{1} \ score$. Simultaneously, the average CPU usage across all configurations remains relatively high, highlighting its efficiency in utilizing available computational resources.

Considering the CPU usage, DFL with a fully connected topology stands at 78 \%. This observation can be linked to the high communication requirements and the corresponding processing needed for maintaining such a topology. In contrast, the RAM usage remains comparably stable among the other configurations, indicating the balanced resource allocation within the DFL architecture. This evidence further supports the notion that Fedstellar is a memory-efficient system with minimal impact on processing requirements from changes in federation architecture.

The SDFL architectures also demonstrate positive results despite a slightly lower $F_{1} \ score$ compared to DFL and CFL. SDFL exhibits efficient resource consumption, reducing network usage to approximately 1148 MB, CPU usage to 69 \%, and RAM to 28 \% when compared to the corresponding values under DFL. These numbers indicate an excellent balance between resource usage and performance. This balance, coupled with an adequate $F_{1} \ score$, makes SDFL an appealing option, synthesizing the benefits of both centralized and decentralized architectures. When moving to the CIFAR-10 dataset, as seen in \tablename~\ref{tab:metrics-CIFAR10}, there is a noticeable decrease in performance under similar configurations. This drop reflects the increased complexity of the CIFAR-10 dataset compared to MNIST. Despite this, the DFL architecture with fully connected topology remains robust, achieving the highest $F_{1} \ score$ of 91.2 \% within the DFL configuration. Though slightly lower than CFL with 92.4 \%, this score is attained in approximately 33 minutes, about 10 minutes faster than CFL.

This quick convergence highlights the adaptability of the proposed platform with more complex datasets. The increase in CPU usage of 80 \% and network usage of 1280 MB can be attributed to the complexity and size of the CIFAR-10 dataset, which has color images compared to grayscale images of MNIST. Furthermore, the $F_{1} \ scores$ for SDFL configurations indicate a small decrease ranging from 97.3 \% to 89.5 \%, which could be improved with more robust selection techniques or using multiple aggregators simultaneously.

In order to demonstrate the scalability of Fedstellar, additional experiments were conducted with 10, 20, 50, and 100 virtualized participants. In this sense, DFL with MNIST dataset and a fully connected network topology were utilized, renowned for their superior performance in previous experiments. As depicted in \tablename~\ref{tab:scalability}, the results show a clear enhancement trend in the $F_{1} \ score$, especially between ten and twenty participants, which moves from 0.975 to 0.987. Beyond this point, the score continues to improve, although the increments are more modest. Regarding resource utilization, CPU usage gradually increases, aligning with the rise in participant numbers. Contrary, RAM usage spikes at ten participants and then tends to decrease, reflecting the diminishing size of the local dataset as more participants join the network. The network usage understandably grows with the number of participants, given the increased communication required in a fully connected topology. Notably, the time to reach an $F_{1} \ score \geq 90\ \%$ significantly drops from approximately 35 minutes with ten participants to around 19 minutes with one hundred participants.

\begin{table}[ht!]
\caption{Scalability of Fedstellar using MNIST dataset with a fully connected topology}
\label{tab:scalability}
\centering
\resizebox{\columnwidth}{!}{
\begin{threeparttable}
\begin{tabular}{c c c c c c}
\hline
\textbf{Participants} & \makecell[r]{\textbf{Model}\\$(F_{1} \ score)$} & \makecell[r]{\textbf{CPU}\\$(\%)$} & \makecell[r]{\textbf{RAM}\\$(\%)$} & \makecell[r]{\textbf{Network}\\$(\text{MB})$} & \makecell[r]{\textbf{Time*}\\$(min.)$} \\
\hline
10 & $0.975 \pm 0.015$ & $73 \pm 12$ & $41 \pm 5$ & $\approx 950$ & $\approx 35$ \\
20 & $0.987 \pm 0.009$ & $77 \pm 15$ & $29 \pm 6$ & $\approx 1243$ & $\approx 28$ \\
50 & $0.991 \pm 0.011$ & $79 \pm 15$ & $30 \pm 8$ & $\approx 2100$ & $\approx 21$ \\
100 & $0.993 \pm 0.004$ & $80 \pm 12$ & $28 \pm 7$ & $\approx 3200$ & $\approx 19$ \\
\hline
\end{tabular}
\begin{tablenotes}
\item * Overall time to reach model $F_{1} \ score \geq 90\ \%$
\end{tablenotes}
\end{threeparttable}}
\end{table}

In conclusion, the results from the virtualized scenario underscore the versatile capabilities of Fedstellar across diverse federated architectures, network topologies, and dataset complexities. While CFL demonstrates impressive performance, in conjunction with a fully connected topology, DFL offers a compelling blend of efficiency, rapid convergence, and robust performance, even when dealing with the more complex datasets like CIFAR-10. The scalability experiments, extending to federations of 100 participants, further validate its efficiency and reliability. This demonstrates the platform readiness to handle a variety of AI tasks effectively, making it a promising tool for flexible and efficient FL applications.

\subsection{Discussion}

The landscape of FL is replete with diverse solutions, each bringing its unique strengths and challenges. The analysis presented in \tablename~\ref{tab:compare-performance} offers a comparative perspective on the performance metrics of Fedstellar against other prominent open-source solutions, as outlined in \tablename~\ref{tab:compare-funcionality}. For a fair comparison, the analysis proposed topologies with twenty virtual participants. CFL solutions underwent evaluation using a star topology, while DFL solutions faced assessment using a fully connected topology. All evaluations are based on the LeNet5 model and MNIST dataset, maintaining a consistent model and aggregation algorithm as with Fedstellar (see Section \ref{sec:results-virtual}).

\begin{table}[ht!]
\caption{Comparison of performance provided by open-source solutions} \label{tab:compare-performance}
\resizebox{\columnwidth}{!}{
\centering
\begin{threeparttable}
\begin{tabular}{c c c c c c c}
\hline
\textbf{Solution} & \textbf{\makecell{Federation\\Architecture}} & \makecell[r]{\textbf{Model}\\$(F_{1} \ score)$} & \makecell[r]{\textbf{CPU}\\$(\%)$} & \makecell[r]{\textbf{RAM}\\$(\%)$} & \makecell[r]{\textbf{Network}\\$(\text{MB})$} & \makecell[r]{\textbf{Time*}\\$(min.)$} \\
\hline
TFF \citep{google:tff:2019} & CFL & 0.982 & 62 & 30 & 1100 & 43 \\
\hline
FedML \citep{He:fedml:2020} & CFL & 0.989 & 59 & 28 & 1050 & 41 \\
\hline
FATE \citep{fedai:fate:2021} & CFL & 0.984 & 58 & 29 & 1091 & 41 \\
\hline
Scatterbrained \citep{Wilt:scatterbrained:2021} & DFL & 0.918 & 89 & 37 & 2631 & 45 \\
\hline
\multirow{2}{*}{\textbf{Fedstellar} (This work)} & CFL & 0.992 & 58 & 26 & 985 & 40 \\
& DFL & 0.987 & 78 & 29 & 1243 & 28 \\
\hline
\end{tabular}
\begin{tablenotes}
\item * Overall time to reach model $F_{1} \ score \geq 90\ \%$
\end{tablenotes}
\end{threeparttable}}
\end{table}

Regarding model performance, most frameworks achieved remarkable results, with $F_{1} \ scores$ around 98\%. However, the CFL configuration of Fedstellar slightly surpassed its counterparts with a score of 99.2\%, potentially due to its implementation of thread lockers ensuring meticulous synchronization with the server and efficient management during aggregation. Scatterbrained, a relevant DFL solution, achieved 91.8\%, while Fedstellar reached 98.7\%, likely a result of its efficient use of message queues for aggregation management and asynchronous communication capabilities. In evaluating resource usage, TFF, FATE, and FedML maintained CPU usage at around 60\%, while Scatterbrained required 89\%. Fedstellar improved CPU usage in CFL and DFL approaches, reflecting its streamlined computational protocols in aggregations and adaptive waiting time between contributions to federated models. When considering network consumption, Scatterbrained utilized 2631 MB, while Fedstellar maintained a consumption of 1243 MB. This efficiency in network usage by Fedstellar and other CFL solutions is largely due to their implementation of adaptive gossip algorithms, effectively reducing the volume of communication required. Training time differentiates these solutions further. While most solutions ranged between 41 and 43 minutes, Fedstellar with DFL configuration significantly reduced the training duration to 28 minutes, owing to its multithreaded computational capability and reduced system latency and bandwidth in inter-participant exchanges.
\section{Conclusion}\label{sec:conclusion}

\change{This work presents Fedstellar, an innovative platform enabling the decentralized training of FL models across diverse physical and virtualized devices.}{This work presents Fedstellar, a platform extension of a previous version proposed by the p2pfl library, which enables the decentralized training of FL models across diverse physical and virtualized devices.} It facilitates the creation and management of DFL, SDFL, and CFL architectures, deployment of complex network topologies, and usage of assorted ML/DL models and datasets, all while enabling user-friendly customization and efficient training process monitoring. The platform incorporates a modular architecture comprising an intuitive frontend for experiment setup and monitoring, a controller for effective operations orchestration, and a core component deployed in each device for federated training and communication. This design makes Fedstellar a comprehensive solution for diverse network topologies and IoT scenarios. Fedstellar was tested across two federated scenarios: (i) a physical one, composed of eight resource-constrained devices, such as Raspberry Pis and Rock64s, acting as spectrum sensors subjected to cyberattacks, and (ii) a virtualized scenario dealing with twenty Docker containers using the well-known MNIST and CIFAR-10 datasets. These tests demonstrated their flexibility and performance, with $F_{1} \ score$ of 91\% in the physical deployment using DFL with a fully connected topology, while 98\% and 97.3\% were obtained using DFL and SDFL with the MNIST dataset. Similarly, the SDFL architecture showcased an approximate 8\% reduction in network usage, a slight trade-off for federated model performance, thus underscoring the balance between resource optimization and communication efficiency.

As future work, there remains a multitude of avenues for exploration. Further research could expand these evaluations to various data types and network topologies. Moreover, additional investigation into the specific performance dynamics of DFL and SDFL could be insightful, particularly in developing techniques for improving its $F_{1} \ score$ without significantly increasing network overhead. Other open research challenges include optimizing the federated training process, exploring more robust secure techniques, and using multiple simultaneous aggregators. In mobility and network dynamics, further investigation could focus on optimizing neighbor selection based on real-time network conditions, such as latency and bandwidth, as well as implementing strategies to identify and isolate malicious nodes that aim to compromise the federated model. Additionally, future iterations of the platform could also explore dynamic timeout strategies, adapting in real time to the network conditions and the load on aggregators. By addressing these challenges, further progress can be made in advancing the performance and adaptability of Fedstellar.

\section*{Acknowledgments}

\addtxt{The authors would like to extend their sincere appreciation to everyone who contributed to the development and support of this research. We extend special acknowledgment to Pedro Guijas and their supervisors, Daniel Rivero and Enrique Fernandez, whose open-source framework provided an indispensable foundation for our platform across base federated models and P2P communications.}

\section*{Funding}
This work has been partially supported by \textit{(a)} 21629/FPI/21, Fundación Séneca, Región de Murcia (Spain), \textit{(b)} the strategic project DEFENDER from the Spanish National Institute of Cybersecurity (INCIBE) and by the Recovery, Transformation and Resilience Plan, Next Generation EU, \textit{(c)} the Swiss Federal Office for Defense Procurement (armasuisse) with the DEFENDIS and CyberForce projects (CYD-C-2020003), \textit{(d)} the University of Zürich UZH, \textit{(e)} MCIN/AEI/10.13039/501100011033, NextGenerationEU/PRTR, UE, under grant TED2021-129300B-I00, and \textit{(f)} MCIN/AEI/ 10.13039/501100011033/FEDER, UE, under grant PID2021-122466OB-I00.

\bibliography{references}

\end{document}